\documentclass[paper]{elsarticle}

\usepackage{lineno,hyperref}
\begin{document}

\begin{frontmatter}

\title{A Distributed Extension of the Turing Machine}
%\tnotetext[mytitlenote]{Fully documented templates are available in the elsarticle package on \href{http://www.ctan.org/tex-archive/macros/latex/contrib/elsarticle}{CTAN}.}

%% Group authors per affiliation:
\author{Luis A. Pineda\fnref{myfootnote}}
\address{Universidad Nacional Aut\'onoma de M\'exico}
\fntext[myfootnote]{lpineda@unam.mx}

%% or include affiliations in footnotes:
%\author[mymainaddress,mysecondaryaddress]{Elsevier Inc}
%\ead[url]{www.elsevier.com}

%\author[mysecondaryaddress]{Global Customer Service\corref{mycorrespondingauthor}}
%\cortext[mycorrespondingauthor]{Corresponding author}
%\ead{support@elsevier.com}

%\address[mymainaddress]{1600 John F Kennedy Boulevard, Philadelphia}
%\address[mysecondaryaddress]{360 Park Avenue South, New York}

\begin{abstract}
The Turing Machine has two implicit properties that depend on its underlying notion of computing: the format is fully determinate and computations are information preserving. Distributed representations lack these properties and cannot be fully captured by Turing's standard model. To address this limitation a distributed extension of the Turing Machine is introduced in this paper. 
In the extended machine, functions and abstractions are expressed extensionally and computations are entropic.
The machine is applied to the definition of an associative memory, with its corresponding memory register, recognition and retrieval operations. 
The memory is tested with an experiment for storing and recognizing hand written digits with satisfactory results. 
The experiment can be seen as a proof of concept that information can be stored and processed effectively in a highly distributed fashion 
using a symbolic but not fully determinate format.
The new machine augments the symbolic mode of computing with consequences on the way Church Thesis is understood. The paper is concluded with a discussion of some implications of the extended machine for Artificial Intelligence and Cognition.
\end{abstract}


\end{frontmatter}

\newtheorem{mydef}{Definition}
%\linenumbers

\section{Distributed versus Turing Machine Computations}

The Turing Machine (TM) and its associated theory of computing provides a fully adequate and widely accepted model of information processing machines.\footnote{For a formal but intuitive discussion of Turing's model and the aspects of the theory of computing addressed in this paper see \cite{Boolos-Jeffrey}.}  According to the strongest version of Church Thesis all general enough alternative models are equivalent to Turing's model which, for this reason, is the most powerful model of computing in every possible sense. Despite the time that has passed since its formulation, this thesis is still held as true by the vast majority of computer scientists. 

However, this claim has been questioned from the perspective of Connectionism and Artificial Neural Networks (ANNs). This paradigm sustains that intelligence emerges from a large number of simple processing units that interact locally and that representations are distributed over a large set of such units. Connectionism has also claimed that Turing's model is limited for capturing appropriately this kind of computations and cannot model adequately cognitive processes like perception, memory and language, among many other mental and motor processes exhibited by biological entities. For instance see the preface of Parallel Distributed Processing (PDP) \cite{Rumelhart} in which this claim was made explicitly. Similar claims are made today within the so-called Embodied Cognition \cite{Anderson-b:2003} and the enactivist approach to Cognition \cite{Froese}. 

The nature of distributed representations is explained by Hinton et al. \cite{Hinton-1986} in terms of the relation between the basic computing or memory units at the level of the hardware and the entities they represent: if this relation is one-to-one the representation is called \emph{local}. In distributed representations, on their part, this relation is many-to-many. An entity may be large and may need be represented by a set of memory units, but the essential aspect of the distinction is that in local representations the memory allocated to represent a basic unit of content --a basic concept-- is independent of the memory allocated to represent all other basic entities, while in distributed representations these allocations may overlap in arbitrary ways.

A side effect of these relations is that entities that have never been input can nevertheless be represented if they are similar enough to the entities already in the representation, so distributed representations can form similarity classes naturally by their inner operations and distributed systems have the capability to \emph{generalize} \cite{Hinton-1986}. These effects cannot be expected in local representations.

Whether a representation is local or distributed has a direct impact on memory. In local representations a particular unit of information is stored in a particular \emph{place} that has an address, which is required to store, read or modify the information. However, human and other biological memories are best thought as distributed representations where contents are distributed over a large number of computing or memory units, that are accessed and retrieved by content \cite{Anderson1980}. For a discussion of computational models  see the introduction of Parallel Models of Associative Memory \cite{Hinton}.

For all these reasons, there is a sense in which Connectionist Systems are indeed more powerful than the standard TM. The challenge to Church Thesis depends on whether the format of the TM is local and distributed systems have for this reason a mode of computing that is essentially different and cannot be captured by the standard TM model. 

However, the claim is obscured in practice because ANNs are normally specified through arithmetic functions which are Turing computable and are simulated in standard computers, with standard data structures and algorithms, using standard programming languages, and these programs are not different in kind to other standard computer programs. Hence, computations are no longer performed by simple units that interact locally. Accordingly, either the TM do support distributed computations after all or distributed computations reduce to local ones. Furthermore, all TMs can be simulated by neural networks \cite{Siegelmann}, and as ANNs are specified as computable functions, these formalisms are equivalent in this sense too.

Despite this, the intuition that there is a fundamental opposition between local and distributed representations and computations is very strong, and whether this is a genuine distinction has implications on the way Church Thesis is understood and, more generally, on the underlying notion or concept on which theories of computing are built up.

In this paper these issues are explored further. In order to ground the discussion and to establish the terminology, the structure and main properties of the standard Turing Machine are presented in Section \ref{properties-TM}. Two underlying properties of the machine, that are called here \emph{determinacy of format} and \emph{information conservation}, are presented and discussed. The determinacy of the format corresponds to the local property of non-distributed representations, but abstracted away from Connectionism and ANNs connotations. So, the format of the TM is indeed local. These two properties taken together involve a third negative property of TMs: these machines are entropy free. The entropy is central to Shannon's information theory, and its absence from the standard theory of computing is somehow paradoxical as computers are the paradigmatic information processing machines.

The main claims of this paper are presented next in Section \ref{distributed-machine}: that there are symbolic computing formats that are not determinate neither information preserving; that distributed representations and machines are characterized precisely by the use of non-determinate formats; that non-determinate formats are entropic and a notion of computational entropy is required, and that distributed extensions of the standard TM using non-determinate and entropic formats can be defined.
An instance of such a kind of machine is presented. In this machine functions and abstractions are expressed extensionally and, unlike the standard TM in which the algorithms codified in the transition table are fixed along the computation, the functions and abstractions computed by the extended machine can be changed by its operations, and the machine is interactive.

An application of the machine to the definition of an associative memory --which uses a highly distributed representation-- is presented in Section \ref{associative-mem}. This memory is framed independently of the connectionist paradigm \cite{Hinton} and also from symbolic associative memories, like semantic networks \cite{Quillian}. Then an experiment on visual memory using the associative memory is presented, where the represented objects, in this case the digits from \emph{0} to \emph{9}, are registered and recognized with very high precision and recall by a fully parallel and highly distributed entropic process. The experiment can be seen as a proof of concept of the possibility of expressing and computing information in a highly distributed fashion at the symbol level directly. 

The implications of the distributed extension of the Turing Machine for the theory of computing and Church Thesis are discussed in Section \ref{implications-for-computing}. This discussion is started with the notion of \emph{mode of computing} which is introduced in order to distinguish TMs, which are symbolic, from other kinds of machines whose mode of computing is not symbolic. A fundamental trade-off between distributed and TM computations is also presented and discussed in this section. This trade-off consists in that distributed computations favor expressivity of extensional information with very effective computation but at the cost of information loss, while determinate computations, where information is expressed intensionally, favor precision and information conservation but have to face the underlying trade-off between expressivity and tractability. This is also a trade-off between natural computations, which are distributed, and artificial computations characterized by the TM model. 

The implications of the distributed extension of the TM for Artificial Intelligence and Cognition, including the opposition between ANNs and TMs, and also between \emph{natural} versus \emph{artificial} computations are discussed in the conclusion in Section \ref{implications-for-cognition}.

\section{Structure and Properties of the Turing Machine}
\label{properties-TM}

The TM is a device for computing functions by mapping their arguments into their corresponding values through the
execution of a fully mechanical procedure, without human intervention. The constituents of the machine are:

\begin{enumerate}
\item A finite set of states $Q$ including a designated initial state $q_{0}$.
\item A finite set of symbols $\Sigma$  --the alphabet.
\item A tape with an unlimited amount of cells --can be extended with no limits, say by its right side-- where each cell holds an instance of a symbol of $\Sigma$.
\item A finite state control with a local scanner device that is above or inspects a particular cell at every state.
\item A set of operations $O = \{R, W_{s_{i} \in \Sigma}, M_{r}, M_{l}\}$  --read, write symbol $s$ for all symbols in the alphabet $\Sigma$, move right and move left-- that are performed by the scanning device at any given state.
\item A transition function --or transition table-- $\delta: Q \times \Sigma \rightarrow (O, Q)$ mapping a state and a symbol behind the scanning device into an operation and the next state of the machine at each computation step.
\item A set of final states $F \subseteq Q$ --some of the states are final.
\end{enumerate}

\begin{figure}
\includegraphics[width=0.5\textwidth]{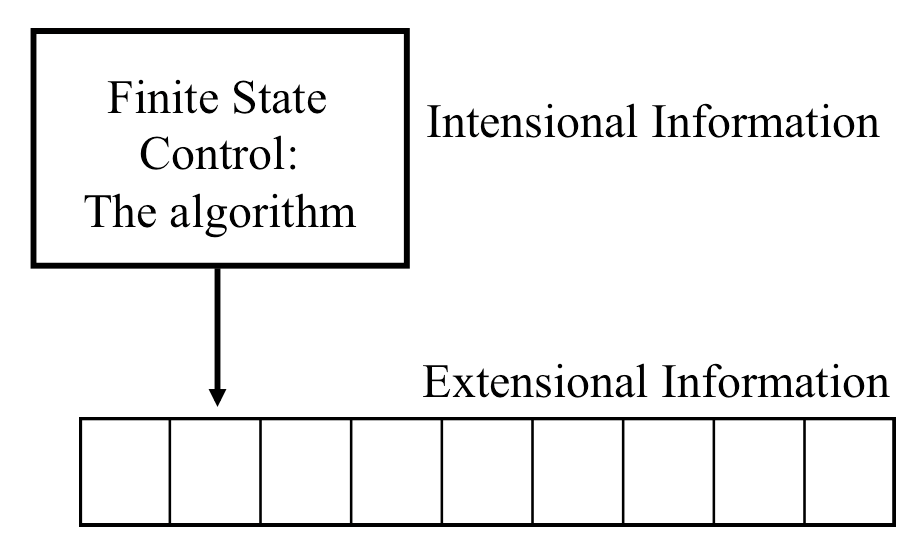}
\centering
\caption{Turing Machine}
\label{Turing-Machine}
\end{figure}

Algorithms --i.e., the mechanical procedures for finding the values corresponding to the given argument-- are codified in the transition function. Algorithms can also be considered as descriptions of their corresponding functions, and the information codified within the finite state control constitutes intensional information. Strings of symbols on the tape, on their part, represent particular arguments and values, and constitute extensional information. The machine is illustrated in Figure \ref{Turing-Machine}.

\begin{figure}
\includegraphics[width=0.7\textwidth]{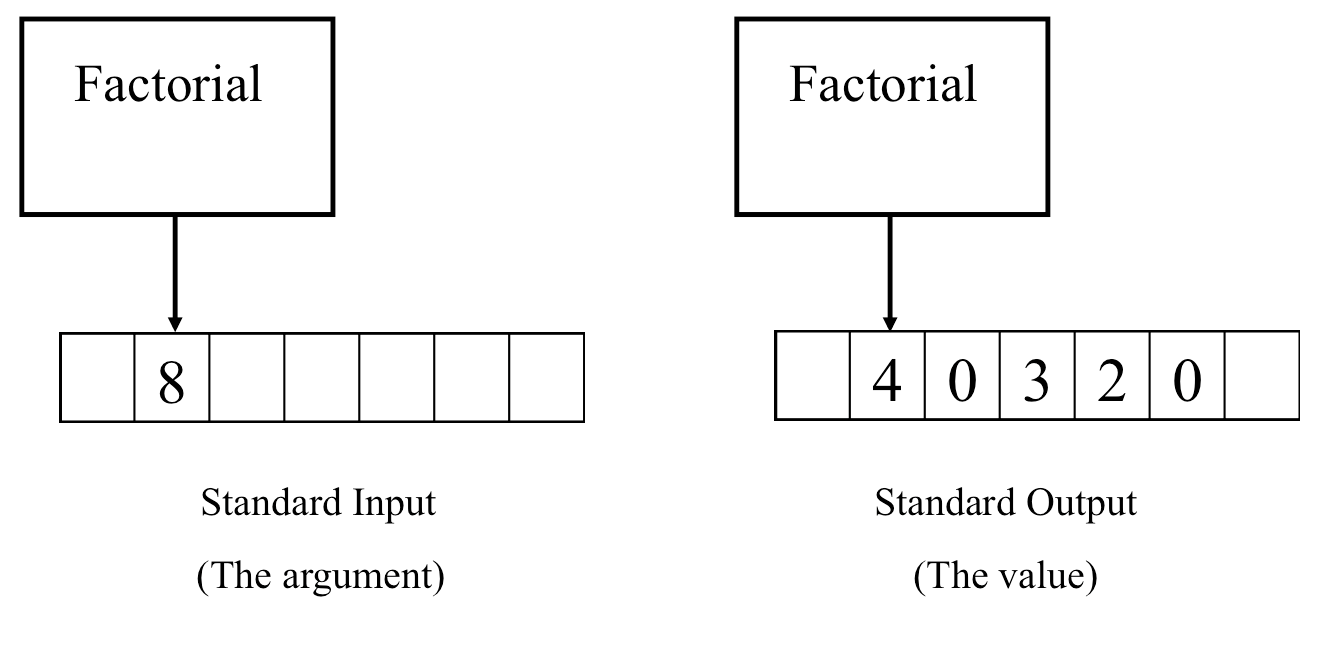}
\centering
\caption{Standard Configurations}
\label{standard-config}
\end{figure}

Computations are interpreted in relation to a predefined set of conventions --including the notation-- that gives rise to the notion of \emph{standard configuration}. In this, the scanning device is at the left-most symbol of the argument at the initial state, and at the left-most symbol of the value at the final state, as illustrated for a particular function  --in decimal notation-- in Figure \ref{standard-config}. 

The TM model does not impose the restriction that the next state for a given state and symbol that is currently behind the scanner should be unique, and when there are more than one  the machine is said to be \emph{non-deterministic}. Computations in these machines unfold as non-sequential structures (e.g. like trees) which can be explored sequentially by the same actual machine or by a set of machines working in parallel where each machine explores a particular path, but this is a contingency due to idealizations of the abstract model, as will be seen below.

The TM has not an explicit theory of memory but information is placed in two kinds of places: 1) the physical substratum in which the transition function is represented and 2) the tape. The theory is neutral in relation to the nature of the former but it can be thought of as implemented in the hardware directly (as it was the case for the first digital computers) or as an internal data structure interpreted by a basic procedure or algorithm that directs the operations of the machine, and its structure is commonly represented as a table or a directed graph --where edges are labeled with symbols and operations that map one state into the next. This representation determines the particular machine and is fixed along the computation. The symbols manipulated by the machine, on their part, are placed on the tape, which is thought of as the memory proper for this reason. The transition function can be codified on the tape, as was proposed by Turing in the original presentation of the machine \cite{TuringMachine}, and can be uploaded by a particular algorithm into the state control, and this machine can in principle compute all functions, given rise to \emph{the Universal Computing Machine}. This notion was later made practical with the notion of \emph{stored program} of current computer architectures, so the two kind of places for storing information are similar except that the memory for storing the transition function is finite, as all computer programs are finite strings of symbols.

Individual cells hold instance of symbols and sequences of cells store strings which are the basic units form --here are called ``words"-- that represent individual concepts --the basic units of content. For instance, the strings representing arguments and values of functions are words --the numerals-- that represent the corresponding numbers --the abstract objects. There are then three levels involving the memory that need to be clearly distinguished:

\begin{enumerate}
 \item The cells or memory units constitute the medium of the representation, which is a physical hardware substratum.
 \item The strings of symbols held on such medium which are the formal objects manipulated by the machine.
 \item The units of content or basic concepts that are the meanings assigned to strings or words by human interpreters.
 \end{enumerate}

Strings are conventionally separated by the blank character (although this typographic convention is contingent), and the tape can hold several words at the same time, so composite units of form, like sentences, are interpreted as the corresponding composite units of content --the propositions expressed by the sentences. This gives rise to an alternative standard configuration in which the tape has a sequence of words at the initial state and the machine halts when the whole of the string has been inspected and a final accepting state has been reached depending on whether the input belongs to a given language.
In this configuration the TM computes a characteristic function (i.e., set membership) and there is a TM for every formal language.\footnote{For a classical presentation see  \cite{Hopcroft}.}
This perspective highlights that the format of the Turing Machine is \emph{linguistic}.

There are three very important assumptions of the model:

\begin{enumerate}
\item Cells on the tape can be added as needed, which means that memory is unlimited.
\item Computations are performed instantly, so there are no limitations in computer power.
\item The machine never malfunctions. 
\end{enumerate}

For these reasons the Turing Machine is an abstract model independent of the construction and physical properties of actual machines. In particular, the TM's model is neutral to whether computations are deterministic or non-deterministic and to how many steps or how much memory is required to complete the computation, which are questions related to the complexity of algorithms, but not to whether functions can be computed effectively or intuitively.

The Turing Machine provides a mechanism for finding the values of all arguments of all computable functions --which can be computed intuitively by people or machines. There are three cases: 
\begin{enumerate}
\item If the machine halts in the standard configuration the argument has the value represented by the word in the final state. 
\item If the machine halts in a configuration different from the standard one, the value of the function for the given argument is not defined and the function is partial.
\item If the machine does not halt the value of the function for the given argument is not defined and the function is partial too.
\end{enumerate}

Condition 3 requires to know in advance whether the machine will halt for the given argument. This can be known by God, because if the machine does not halt immediately it will never halt at all. However, mortals need a machine to this effect, \emph{the halting machine}, that has as its arguments the particular machine (i.e., its \emph{id} or its description) and the argument in question, and has as its value say 1 if such machine halts for such argument and 2 if it does not. The definition of this machine is known as the \emph{halting problem}. There are machines that never halt and that can be known by inspection of their structure, but it is known that the halting problem for an arbitrary function and argument cannot be solved. The halting problem is an instance of a non-computable function, and there are several functions that are not computable. These are the limits of the theory of computability in the sense of the Turing Machine.

Other general models that use symbolic manipulation as their mode of computation, like the theory of recursive functions, the Von Neumann architecture or \emph{Abacus} computations \cite{Boolos-Jeffrey} and the $\lambda$-calculus, are equivalent to the Turing Machine, in the sense that the description of every TM can be translated or reduced into a description in the alternative theory and vice versa for all well-formed descriptions in these formalisms. Consequently all functions that can be computed in any of these formalisms can be computed in all the others. These results motivated Church Thesis which states that the Turing Machine computes the set of functions that can be computed by any other general enough model of computing, which also corresponds to the set of functions that can be computed intuitively by people (given enough time and paper). Furthermore, ANNs are TMs, as discussed above, and TMs are ANNs \cite{Siegelmann, Sun}, so these formalisms are equivalent, providing additional support to Church Thesis.

\subsection{Determinacy of the Format}

An implicit assumption of Turing's model is that the format of the information that is placed in the tape is fully determinate. This is the restriction that if an instance symbol belongs to an instance word, it cannot belong to any other word in the tape: if $s_i$ $\in$  $w_j$ then $s_i$ $\notin$ $w_k$ for all $k \neq j$. So, any given word uses memory units that are independent of the memory units used by all other words. Instance symbols are like sub-atomic particles that form words, the atomic representational units. Of course, composite units of form, like phrases or sentences, use the cells that are used by their constituent basic units or by other embedded or subsumed composite units. Substrings, like syllables, carry no content by themselves and can be thought of as sub-atomic particles. In case such particles do carry meanings, such substrings become the atoms that are combined in the construction of a composite word, which is no longer a basic unit of form and content for this reason. This very basic convention allows that phrases and sentences are built up by composition but not by superposition (or overlap), and that the meaning of a composite structure is a function of the meaning of its parts and its mode of combination: the principle of compositionality. 

A word with the concept that it expresses can be thought of as a ``capsule" that is independent of all others, and relations between concepts need to be expressed through additional words, like verbs or prepositions. Conversely, a composite structure can be decomposed into its constituent parts, and the determinacy of the format allows analysis (e.g., parsing an string in relation to a given grammar). 

Determinacy also guarantees that erasing a word from the tape or moving it to another location will not affect the integrity of other words.
The independence of the basic units of form and content underlies that memory objects in TM \emph{have a place} and, in particular, 
the metaphor of memory as a ``chest of drawers"  implemented in Random Access Memory (RAM) of standard computer architectures. For all this, TM representations are local as opposed to distributed as discussed above in Section 1.

The determinacy of the format does not conflict with non-deterministic machines and these notions should not be confused. For highly-expressive languages strings may even be ambiguous (i.e., have more than one interpretation), but it is never the case that the same memory cell or instance symbol is shared by more than one word and the format of non-deterministic machines is fully determinate. 

\subsection{Information Conservation}

Another implicit assumption of Turing's model is that the machine never loses information in computations. This is a property of memory and data bases, as what is registered is exactly what is recovered. More generally the property can be appreciated in the architecture of the machine directly: 1) the transition table 
in the control state --that defines the particular machine-- is fixed along the computation 
and 2) every symbol that is written on a cell of the tape is the same that will be read the next time the scanner visits such cell.

\begin{figure}
\includegraphics[width=0.7\textwidth]{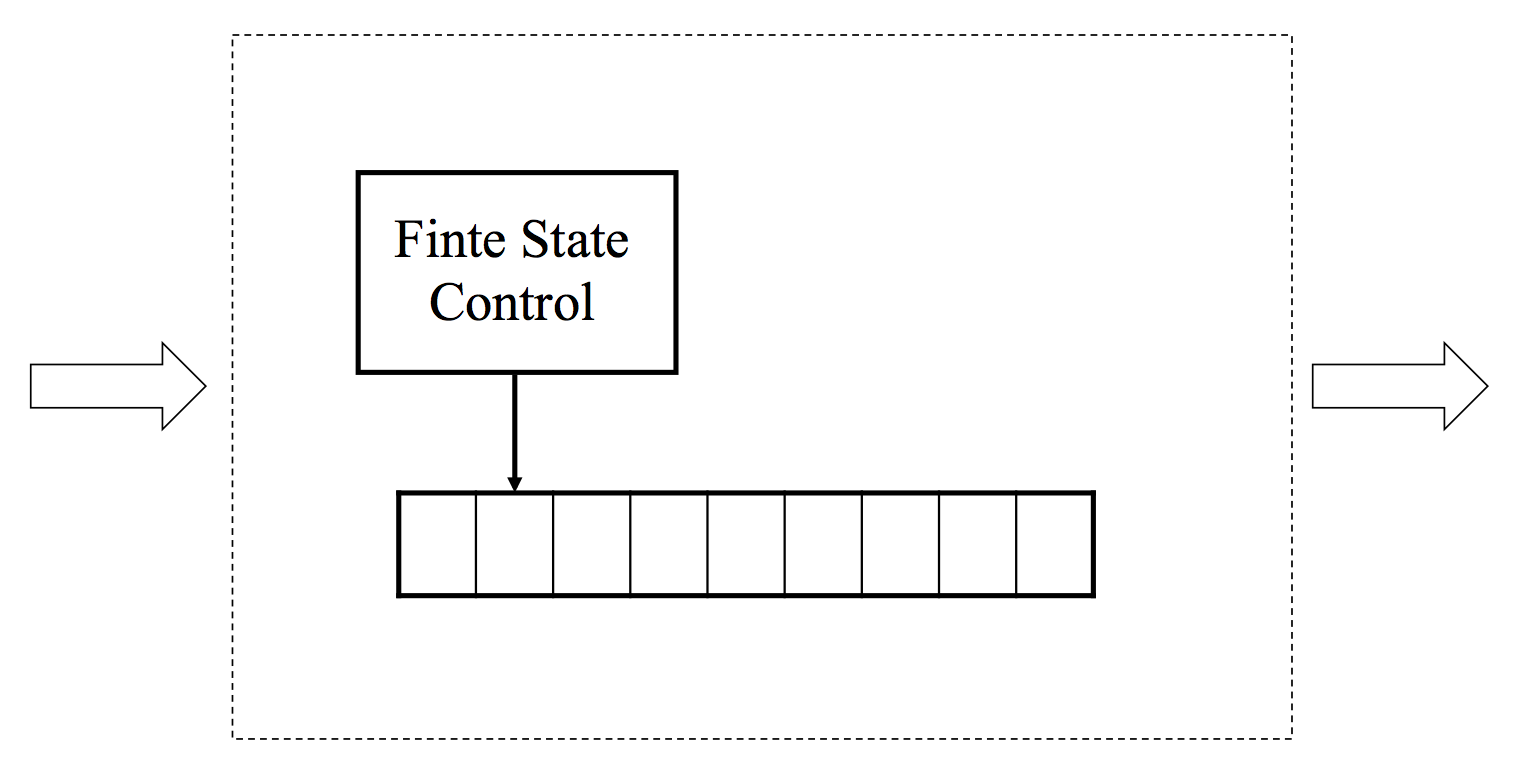}
\centering
\caption{``Control Volume" of TMs}
\label{control-volume}
\end{figure}

Information loss should not be confused with information transformation. 
Writing, modifying or erasing a symbol or a word on the tape is a part of algorithms, and transforming information is the substance of computations. 
In particular, erasing a word consists on writing the blank symbol on the corresponding cells.
In a thermodynamics analogy, information is to energy, computing operations are to work, 
and the Turing Machine as a whole is within an information ``Control Volume", as illustrated in Figure \ref{control-volume}.
In the same way that the Law of Conservation of Energy states that energy is never created or destroyed but only transformed within a control volume,
information is never created or destroyed but only transformed by the Turing Machine.

Information conservation should also be distinguished from \emph{reversibility}. In one sense of this term, computations are reversible --or
invertible-- if $f^{-1}(f(x))=x$, for all $x$.
Reversibility can be achieved by concatenating the machine that computes a function with the one that computes its inverse, which are different machines.
However, some functions have and some lack inverses and not all computations are reversible, but this property and information conservation are different notions.
In another related sense the TM is not reversible because the transition function in its finite state control does not have an inverse,
so from looking at a particular state it is not possible to know the previous one; 
however, if the machine does not have this information in its transition table in the first place, it cannot lose it in computations.
Furthermore, computations can be made reversible in this latter sense too \cite{Bennet, Abramsky}. 
There is another sense of reversibility that refers to the composition and 
decomposition of strings without information loss, which is allowed by 
the determinacy of the format, and reversibility in this sense is information preserving.

\subsection{Computational Entropy}

Determinacy and information conservation underly a negative property of TMs: these machines are entropy free and, consequently, ``entropy" is not a theoretical term in the theory. The entropy is the expected value of the information content of a message in a communication system or a sentence in a language, so computers do not take into account or make use of the amount of information that is carried by the representations that they hold. This should be somehow paradoxical because although computers are the paradigmatic information processing machines and the entropy is the central notion in Shannon's information theory, the entropy is not considered in computations.  The reason is that entropy can also be thought of as the degree of determination of the content of the message or representation. If this is fully determinate the entropy is zero. Functions, the objects of computation, are fully determinate and the representations expressing them are maximally informative. So the entropy of these mathematical objects is zero and this notion is not needed. For the same reason, these representations can never lose information by the actual operations of the machine. Hence, the Turing Machine is a maximally efficient information processing machine, a kind of ``perpetual moving machine". This is another implicit assumption of the mode of computation of the TM and its equivalent formalisms.

This should not be confused with the fact that computers compute the entropy of computer applications, which is done all the time. For instance, computing the entropy of corpora or data sets in natural language or text processing applications is  commonly required. Similarly for machine learning algorithms that use the entropy of the data as a main parameter, like in the construction of decision trees. A large number of application domains involve the entropy, but this should not be confused with the entropy of the representations expressing the knowledge of such domains themselves.

The computational entropy should neither be confused with the physical entropy of analogical computers that use physical modes of computing (e.g., \cite{Siegelmann97analogcomputation}). The claim here is that the symbolic manipulation mode of computing does not involve information entropy, and that for this reason ``entropy" is not a theoretical term in the standard theory of Turing Machines and computability.

\subsection{Final remarks on Turing Machines}

The Turing Machine is an abstract device that is meant as an ideal model of computing that captures the essential aspects of actual computers. In particular the interaction with the world is left to implementations. Once the computation is started the machine takes full control of the tape until the final state is reached, and the information is read or written when the machine is in an ``idle"  state, between the final state of the previous computation and the initial state of the next one, and once a computation is required, a start signal has to be made. The assumption is that other agents, people or machines, can write and read the contents of the tape, which is placed in the input and output standard configurations at the interaction points, and that particular machines can be switched  for performing complex computations at these points too, but these considerations are not part of the abstract model. The need to consider the interaction including unforeseen inputs, the possibility that the machine changes its behavior or evolve over time, and the infinity of operations and continuos input streams, have motivated proposals for extending the basic model \cite{Leeuwen-Wiedemann-2000,Leeuwen-Wiedemann-2001}; however, the need to characterize distributed representations at the functional and computational level seems to be still pending.

\section{Distributed Extension of the Turing Machine}
\label{distributed-machine}

\begin{figure}
\includegraphics[width=0.8\textwidth]{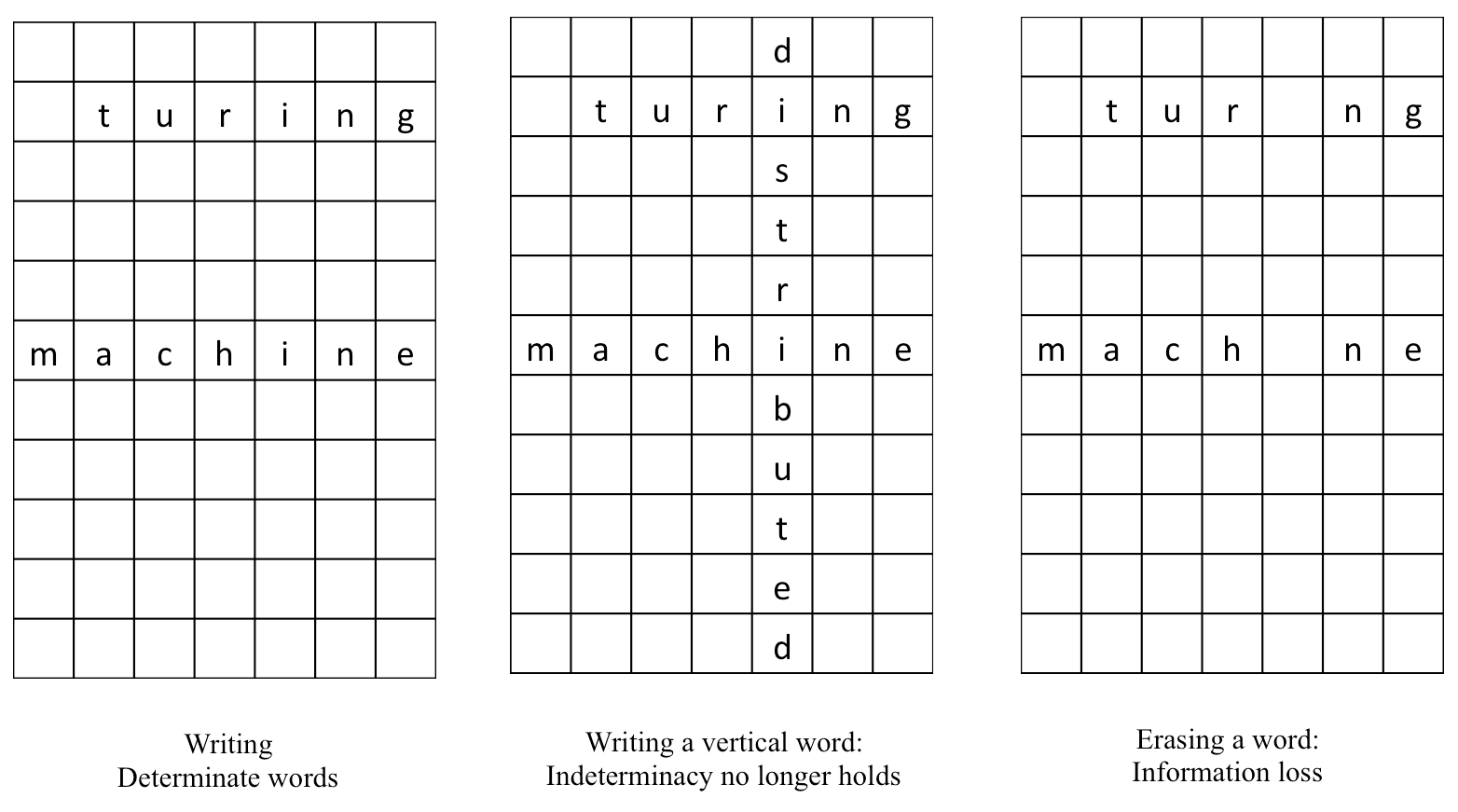}
\centering
\caption{Indeterminacy and information loss}
\label{crossword}
\end{figure}

There are formats that lack the properties of determinacy and information conservation. This is illustrated by the crossword, in which a 2-D grid instead of a tape is used as the representation medium. This format does not satisfy determinacy by definition as the symbols at the intersections belong to more than one word. Furthermore, erasing a word may affect the integrity of the words that intersect with it, and there is information loss. These phenomena are illustrated in Figure \ref{crossword}.

The lack of determinacy and information conservation may be seen as a limitation of the computing format; however, 
whenever a cell of the medium contributes to two or more units of form at the same time, 
a relation between such units --possibly with their associated units of content-- is established at the level of the format. 
This is the basic property of distributed representations. Or more directly, a representation is distributed precisely when its
underlying computing format lacks determinacy. 
Distributed representations capitalize the increased expressive power provided by the shared units, but basic units of form and content are no longer independent. In distributed representations the reversibility of composition and decomposition is limited, and hence the capability of analysis. For all this, the metaphor of storing information in a chest of drawers does not longer apply.

In order to study further computations with an indeterminate format the Turing Machine is extended 
as follows:

\begin{enumerate}
\item The finite state control is extended with an $n$ $\times$ $m$ \emph{Control Grid} $G_{\phi}$.
\item The tape is extended into an $n$ $\times$ $m$ \emph{External Grid} $G_{\psi}$.
\item The scanner is extended to a set of $n$ $\times$ $m$ scanners aligning the corresponding cells of $G_{\phi}$ and $G_{\psi}$.
\item The content of the cells of both $G_{\phi}$ and $G_{\psi}$ are symbols of an alphabet including the mark $X$ and the blank $B$.
\item A set of grid operations \{$\tau$, $\upsilon$,...\} mapping the contents of the cells of $G_{\phi}$ and $G_{\psi}$
from the initial into a final state, such that for all cells $i$, $j$:
\begin{enumerate}
\item $G_{\phi, s_f}(i, j) = \tau(G_{\phi, s_0}(i, j), G_{\psi, s_0}(i, j))$
\item $G_{\psi, s_f}(i, j) = \upsilon(G_{\phi, s_0}(i, j), G_{\psi, s_0}(i, j))$
\end{enumerate}
Similar operations can be defined between cells and columns, and columns and columns.
\end{enumerate}

\begin{figure}
\includegraphics[width=0.5\textwidth]{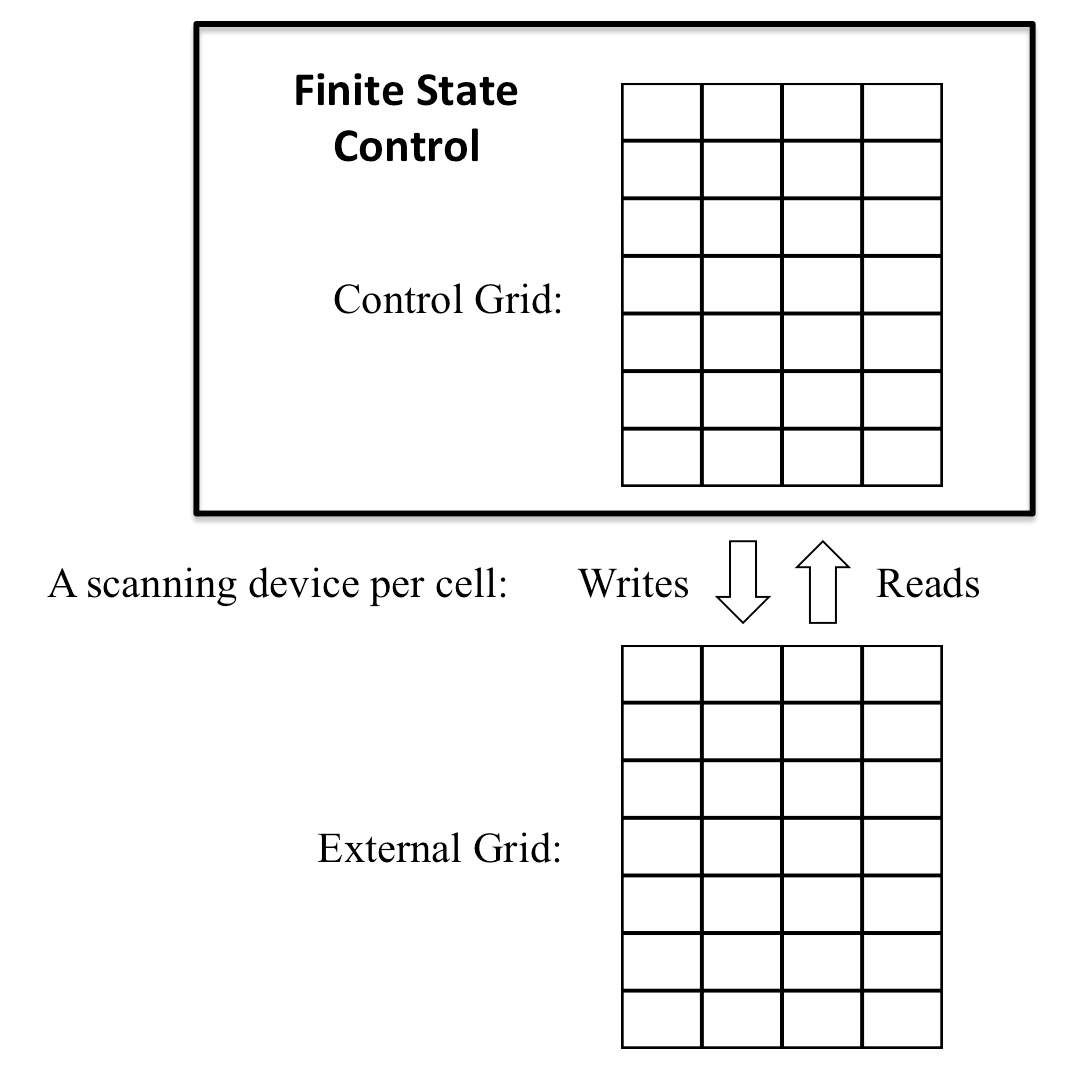}
\centering
\caption{Distributed Computing Machine}
\label{diag-machine}
\end{figure}

The extended machine subsumes the standard one --which does not use the control grid-- and the \emph{move} operations
can be implemented by selecting a particular scanning device locally, so standard algorithms can be defined. Although this functionality is not defined in the present description, it can be added so that a set of scanners can be turned on and off at particular states.
Operations between cells such that the content of a cell of the control grid in the next state is a function of its content in the present state, the content of the corresponding cell in the external grid and the contents of the neighboring cells in the control grid, as in computational retinas \cite{Funt} or diagrammatic processors \cite{Anderson:2003,Anderson:2010}, can also be defined, although this kind of functionality is neither explored further in this paper. 
The extended machine is illustrated in Figure \ref{diag-machine}. 

\begin{figure}
\includegraphics[width=0.2\textwidth]{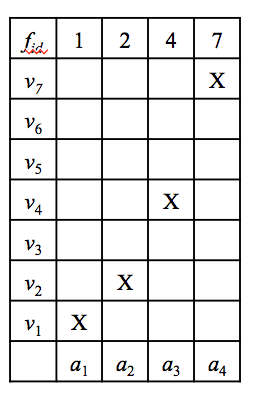}
\centering
\caption{Diagrammatic representation of discrete functions}
\label{dia-grep-func}
\end{figure}

In the basic mode of operation, the control grid $G_{\phi}$ holds a function represented discretely in a table notation instead of the algorithm in the transition table of the TM, as illustrated in Figure \ref{dia-grep-func}. Columns correspond to arguments and rows to values, and a mark on a cell at the intersection between a column and a row represents the value of the corresponding argument. A function is defined if there is at most one mark in each column. The function is defined in the state of the cells, and the arguments, values and the function index in Figure \ref{dia-grep-func} are meta-data for illustration purposes only. 

For the operation of the machine we assume that there is an infinite supply of control and external grids for the specification of finite discrete functions of an arbitrary number of arguments and values. So, there is no memory limitation. As all pairs of $n$ arguments and $m$ values are enumerable, all
total and partial discrete functions are enumerable and can be computed in this format. The enumeration of the functions represented in a grid of size $n \times m$ is as follows:
Let $f_k$ be a function in the enumeration, where the function index $k$ is a numeral 
of $n$ digits in base $m + 1$. The index is composed by the concatenation of the subscripts of the values $v_j$ for all the arguments $a_i$, such that 
$k_i = j$ if $f_k(a_i) = v_j$ and $k_i = 0$ if $f_k(a_i)$ is undefined. 
Total functions  can be identified easily as they have no $0s$ included in their indices. 
The number of total and partial functions is $(m+1)^n$.
The table in Figure \ref{dia-grep-func} illustrates the representation of 
the total function $f_{1247}$ where $n = 4$ and $m = 7$.

\begin{figure}
\includegraphics[width=0.5\textwidth]{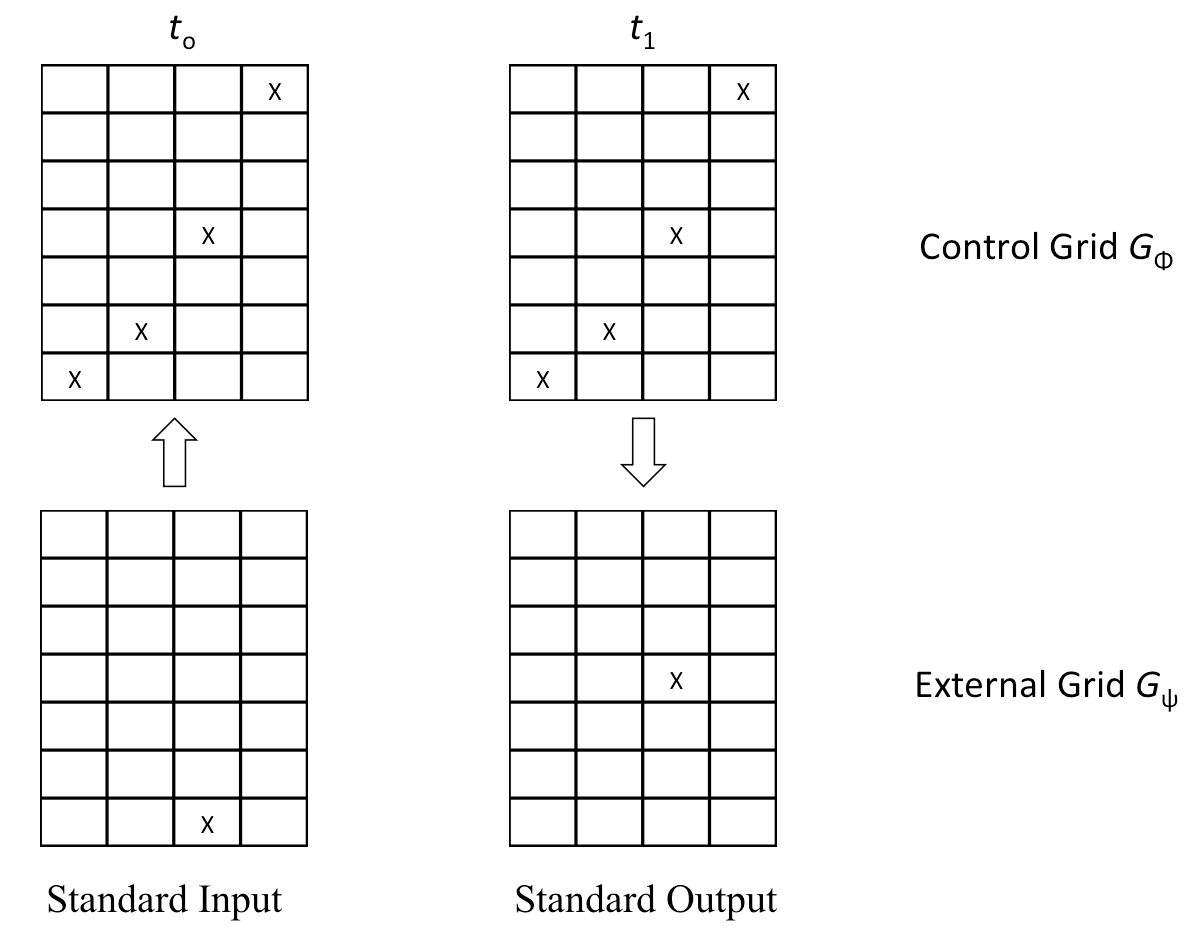}
\centering
\caption{Function Evaluation}
\label{func-eval}
\end{figure}

As in the Turing Machine, functions are evaluated in relation to a standard configuration. For instance, the input and output standard configurations for finding the value of the third argument of function $f_{1247}$ are illustrated in Figure  \ref{func-eval}. The evaluation proceeds by selecting the column of $G_{\phi}$ (up grid) indexed by the mark that corresponds to the argument at the bottom row of $G_{\psi}$ (down grid) in the standard input; erasing the mark in $G_{\psi}$; selecting the mark in the selected column of $G_{\phi}$ and copying such mark back into corresponding cell in $G_{\psi}$ to render the value in the standard output configuration. In this notation all functions are evaluated by this procedure and the format does not require the definition of specific algorithms for particular functions.

\subsection{Distributed Computations}

The potential of the extended machine comes from its capability to support distributed representations that can be computed in a fully parallel way through direct cell-to-cell operations involving a finite and small number of steps. Discrete functions are considered the basic units of form or ``words" and as concepts are represented through functions, these are also the basic units of content. The extensional representation of these objects in both $G_{\phi}$ and $G_{\psi}$ permits the simultaneous representation of more than one word at the same time, like in the crossword. For this, functions interact and the representation becomes distributed. Following the analogy of writing and erasing words in the cross words, we define the corresponding operations that we call \emph{abstraction} and \emph{reduction}. Information is input into  $G_{\phi}$ through $G_{\psi}$. The specification of the operations states the information in  $G_{\psi}$ and $G_{\phi}$ both at the initial and final states.

\subsubsection{Abstraction}

The abstraction operation reads the information in $G_{\psi}$ and overlaps it with the information in $G_{\phi}$. 
In order to support interactivity the external grid  $G_{\psi}$ is reset to blanks at the end of the operation, 
so an external agent can write on it the next function or abstraction to be input directly. 
Also, as part of the standard input configuration $G_{\phi}$ is empty at the start of the computation. The definition of the abstraction operation is:

\begin{mydef}
{\bf Abstraction}
\end{mydef}
For all cells $i$, $j$ and time $k$,
\begin{enumerate}
\item $G_{\phi, t_{k+1}}(i, j) = G_{\phi, t_k}(i, j) \vee G_{\psi, t_k}(i, j)$
\item $G_{\psi, t_{k+1}}(i, j) = 0$
\end{enumerate}

We refer to the contents of these operations as $\Gamma_{\phi} = \lambda_{\phi}(\Gamma_{\phi},\Gamma_{\psi})$
and $\Gamma_{\psi} = \lambda_{\psi}(\Gamma_{\phi},\Gamma_{\psi})$ respectively.
Also, a function $f$ in $G_{\phi}$ is an abstraction $\Gamma_{\phi}$
and a function $g$ in $G_{\psi}$ is an abstraction $\Gamma_{\psi}$.

\begin{figure}
\includegraphics[width=0.8\textwidth]{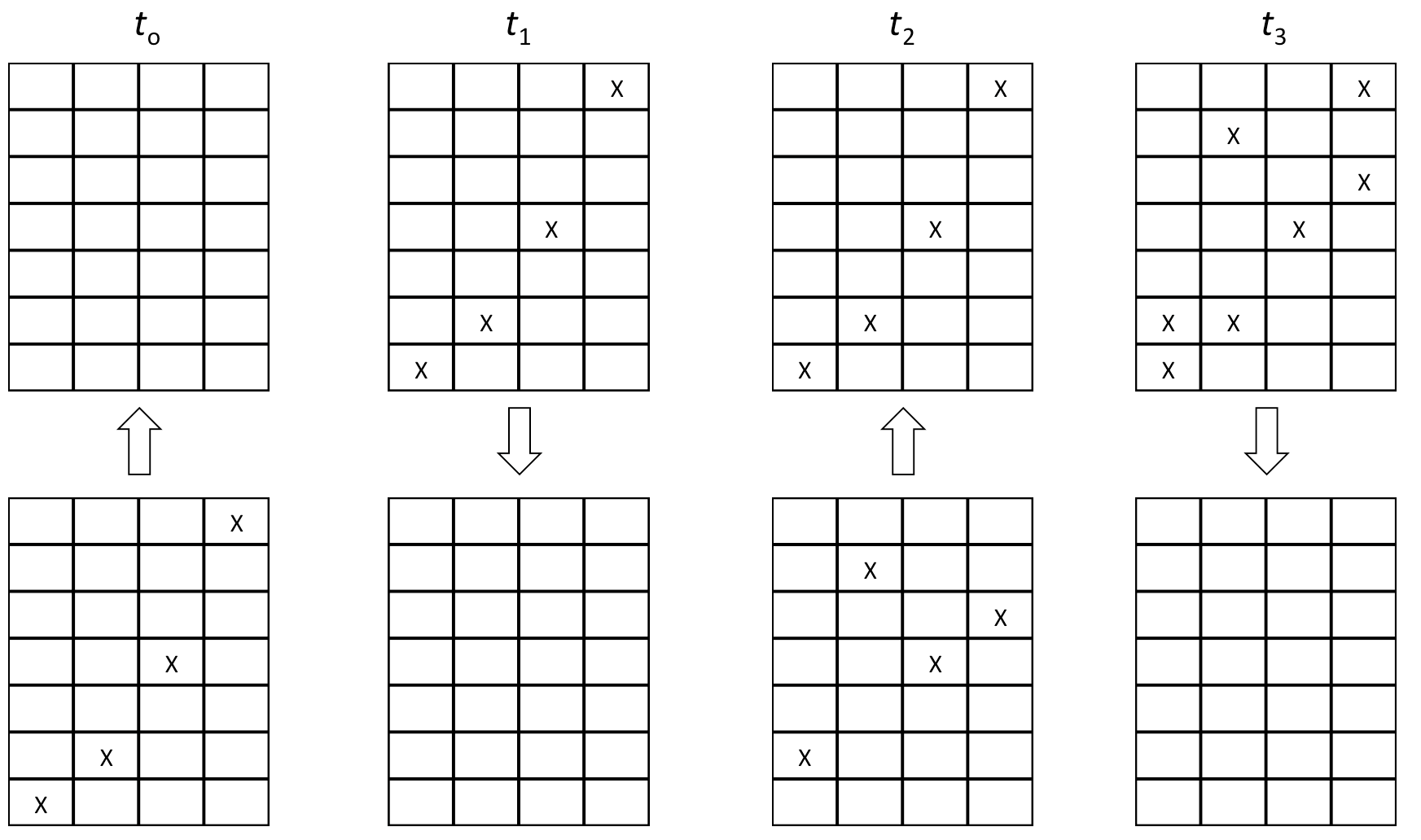}
\centering
\caption{Abstraction}
\label{abstraction}
\end{figure}

The sequential application of two abstraction operations is illustrated in Figure \ref{abstraction}. In the notation a blank space is a \emph{B} in a cell or \emph{0} for logical operations. In this example the content of $G_{\phi}$ is indeterminate when the two functions have been input, as all arguments but the third have two possible values. Indeterminate representations can be used to solve a system of equations by inspection: the solution is the value of the fully determinate argument (i.e., in this case the third one) which corresponds to the intersection of the lines representing the two functions in a graphical representation. The computation is a direct operation on the representation.

Abstractions can be evaluated by a standard procedure, but unlike functions that produce a determinate value, the values of the arguments with more than one mark in the corresponding column are indeterminate. For instance, $f(a_2) = v_2 \vee v_6$ at $t_3$ in Figure \ref{abstraction}. The larger the terms in the disjunction the larger the indeterminacy.

The abstraction operation modifies the function computed by the machine, extending the functionality of the standard TM in this way too. The operation involves also that the machine is interactive in the sense that inputing data from the external grid does impact on the definition of the mathematical object that is been computed, and for the same reason the machine may evolve along the computation. This is possible due to the nature of the extensional information manipulated by the machine, in opposition to the standard TM in which algorithms encapsulate intentional information which cannot be modified along the computation.

The operation also illustrates that the basic object of computation of the extended distributed machine is the abstraction, which is a relation. The function is a particular kind of relation in which all arguments are fully determinate. The relation, on its part, is a non-determinate object as the arguments are associated to a number of values. Functions in distributed representations are like sub-atomic particles or features, that are abstracted upon by superposition in the construction of abstractions, which are the atoms representing individual concepts or units of content.

The operation also shows that in the extended machine the objects of computation are also the memories. The data in the external grid does enrich the content of the object being computed in an abstraction operation, and it is no longer passive data, like the extensional information placed on the tape in the standard TM, and the distinction between programs and data is blurred in the distributed machine.

\subsubsection{Information Enrichment}

\begin{figure}
\includegraphics[width=0.2\textwidth]{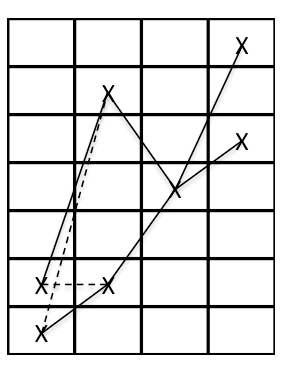}
\centering
\caption{Information enrichment}
\label{inf-enrichment}
\end{figure}

The extension of the set of functions produced by the abstraction operation is augmented with information
due to the geometric properties of the format. This is illustrated in Figure \ref{abstraction} where
$\Gamma_{(1 \vee 2)(2 \vee 6)4(7\vee5)} = \lambda_{\phi}(\Gamma_{1247},\Gamma_{2645})$ contains not two 
but eight functions: all functions that can be formed by taking each possible value for each argument.
Each function corresponds to one of the eight paths that can be formed from the first to the last argument in $G_{\phi}$, as illustrated in Figure \ref{inf-enrichment}.
The segments in solid lines correspond to the overt functions and the ones in dotted lines to the additional information provided by the abstraction.
The additional functions are very similar to the abstracted ones and overlap with them in most of the segments, and 
the enrichment is circumscribed but do provide a generalization space. In an analogy
with inductive learning, a general --but enclosed-- class is generated out of a number of concrete instances;
however the information enrichment is due to the format but not to the content of the instance objects, and differs in this regard 
from machine learning algorithms, including ANN, where the generalization is explained in terms of similarity measures or some kind or another. This effect of information enrichment due to the format has been labeled within diagrammatic reasoning as \emph{free-rides} (e.g., \cite{Shimojima}).

\subsubsection{Reduction}

The reduction operation extracts the function or abstraction in the external grid $G_{\psi}$ out of the control grid $G_{\phi}$. The operation proceeds in two steps: first, a test to check whether $\Gamma_{\psi}$ is included in $\Gamma_{\phi}$ is performed. If the test is satisfied the function or abstraction in $G_{\psi}$ is erased from the abstraction in $G_{\phi}$. 

The inclusion test is defined as a material implication between the abstractions in both grids (i.e., $\Gamma_{\psi} \rightarrow \Gamma_{\phi}$ is true). If the test is satisfied the information content of both cells remain the same. Otherwise, the content of the control grid is not altered but the external grid is set to blanks. This convention indicates that
the test failed and prepares the input buffer for a new test.

\begin{figure}
\includegraphics[width=0.4\textwidth]{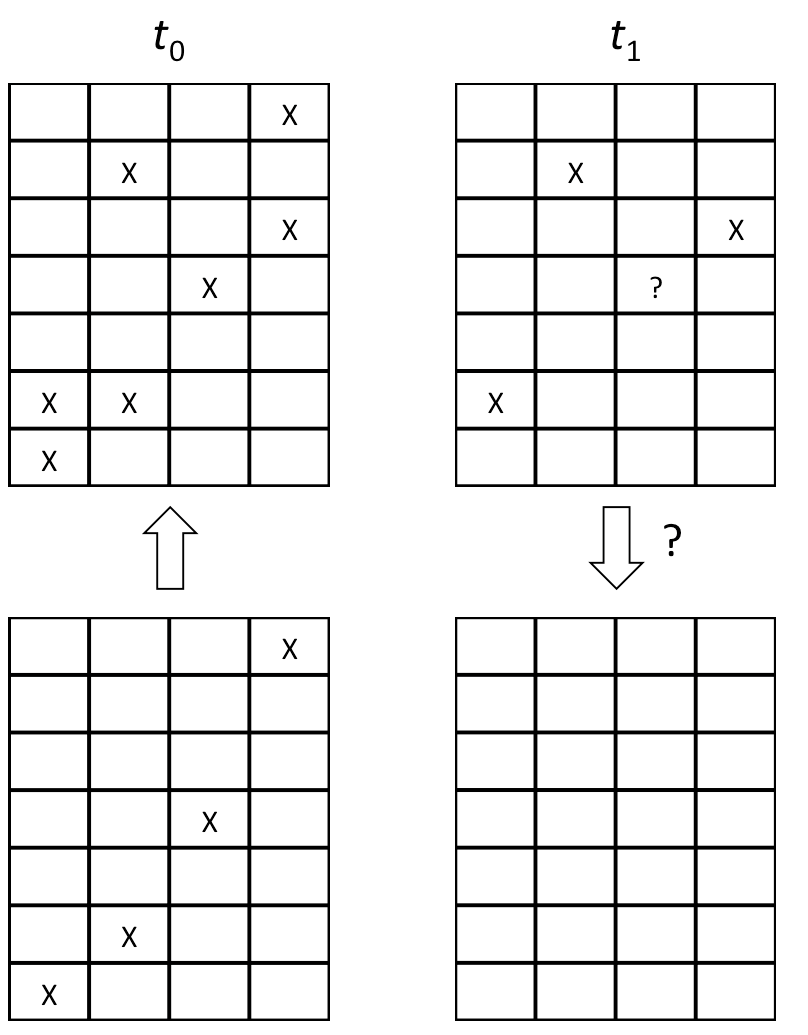}
\centering
\caption{Reduction and indeterminacy}
\label{indeterminacy}
\end{figure}

The second part faces the problems of erasing a word in a crossword: If the functions in $G_{\phi}$ interact, erasing one has a side effect that the other is altered too, and there is information loss. The dillema is illustrated in Figure \ref{indeterminacy} where the function $f_{1247}$ in $G_{\psi}$ should be erased from $G_{\phi}$. 

In general, reduction should be thought of in terms of an optimization problem with two conflicting objectives: 
\begin{enumerate}
\item Extract as much information as possible of the function or abstraction in $G_{\psi}$ out of $G_{\phi}$.
\item The remaining function or abstraction in $G_{\phi}$ should be minimally affected. 
\end{enumerate}

For this particular setting the problem is addressed with a simple heuristics: for each column, if all cells of $G_{\phi}$ and $G_{\psi}$ are equal, it is not known what objects (functions or abstractions) contributed to the construction of $\Gamma_{\phi}$, so leave the column in $G_{\phi}$ as it is; 
otherwise subtract the content of the column of $G_{\psi}$ out of the corresponding column of $G_{\phi}$, as $\Gamma_{\psi}$ must have
contributed to the construction of $\Gamma_{\phi}$. Finally, the content of $G_{\phi}$ is copied out into $G_{\psi}$. That is, $\Gamma_{\psi}$ is a key or an index
to retrieve $\Gamma_{\phi}$. With these considerations reduction is defined as follows: 

\begin{mydef}
{\bf Reduction}
\end{mydef}

\begin{enumerate}
\item{Inclusion test: For all cells $i$, $j$ 
if $G_{\psi, t_k}(i, j) \rightarrow G_{\phi, t_k}(i, j)$ (material implication) is true then
$G_{\psi, t_{k+1}}(i, j) = G_{\psi, t_{k}}(i, j)$; otherwise
$G_{\psi, t_{k+1}}(i, j) = B$.
In any case $G_{\phi, t_{k+1}}(i, j) = G_{\phi, t_{k}}(i, j)$.}

\item{Reduction: For all columns $i$ 
if $G_{\phi, t_k}(i) = G_{\psi, t_k}(i)$ 
then $G_{\phi, t_{k+1}}(i) = G_{\phi, t_k}(i) = G_{\psi, t_k}(i)$
and $G_{\psi, t_{k+1}}(i) = G_{\psi, t_k}(i)$; otherwise $G_{\phi, t_{k+1}}(i) = G_{\phi, t_k}(i) - G_{\psi, t_k}(i)$
(i.e., set difference between columns)
and $G_{\psi, t_{k+1}}(i) = G_{\phi, t_k}(i)$.
We refer to the content of these operations as $\Gamma_{\phi} = \beta_{\phi}(\Gamma_{\phi},\Gamma_{\psi})$
and $\Gamma_{\psi} = \beta_{\psi}(\Gamma_{\phi},\Gamma_{\psi})$ respectively.
}
\end{enumerate}

\begin{figure}
\includegraphics[width=0.4\textwidth]{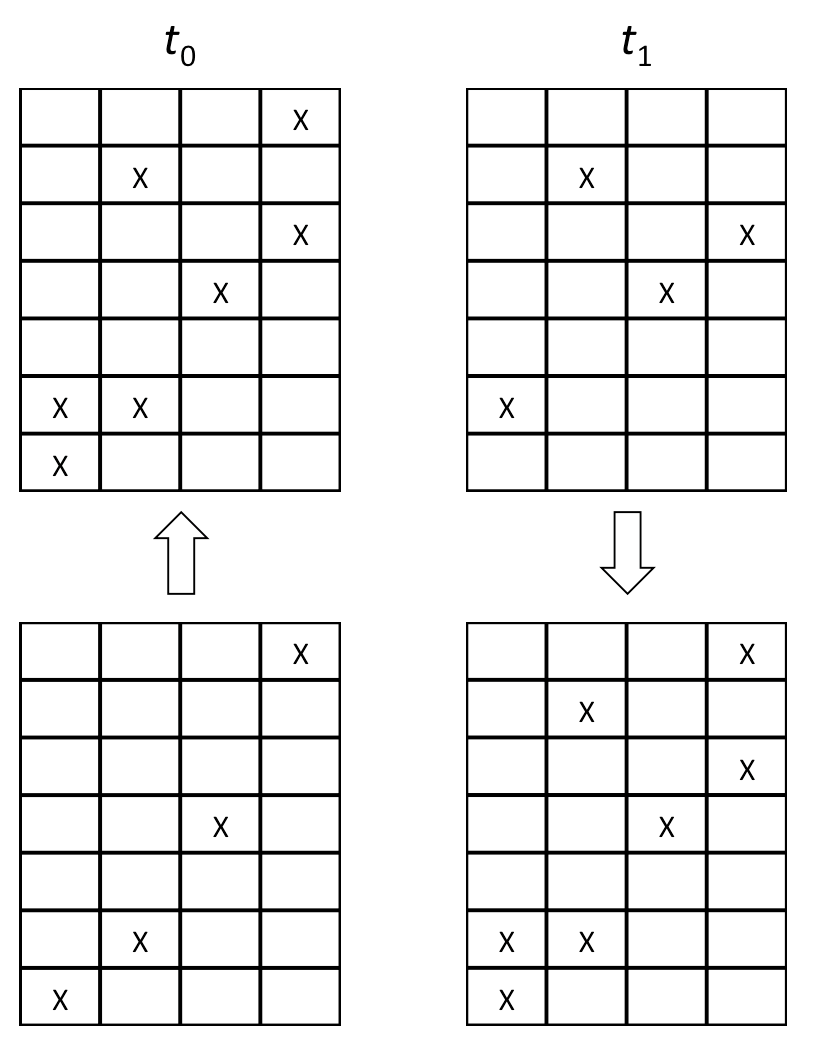}
\centering
\caption{Reduction}
\label{reduction}
\end{figure}

This definition is illustrated in Figure \ref{reduction}.

The indeterminacy of the format has as a consequence that reduction is not information preserving. Information conservation requires that
$\beta_{\phi}$($\lambda_{\phi}$($\Gamma_{\phi}$, $\Gamma_{\psi}$), $\Gamma_{\psi}$) = $\Gamma_{\phi}$ for all $\Gamma_{\phi}$ and $\Gamma_{\psi}$
always holds.
However this is not the case in general as illustrated by the following example:
$\beta_{\phi}$($\lambda_{\phi}$($\Gamma_{2\vee4}$, $\Gamma_{2}$), $\Gamma_{2}$) = 
$\beta_{\phi}$($\Gamma_{2\vee4}$, $\Gamma_{2}$) =
$\Gamma_{4}$ 
but
$\Gamma_{2\vee4} \neq \Gamma_{4}$. Hence, reduction involves information loss.

The distributed machine also differs from the standard TM in that reduction does modify the abstraction or function that is computed, and the distributed machine is interactive and may evolve due to this latter operation too. In the metaphor of memory as a chest of drawers, a drawer may be selected and its content can be read, leaving the drawer intact, but if the drawer is needed to store a different content, it has to be erased without affecting the contents of other drawers. This is what cannot be guaranteed in distributed systems, where information extraction becomes a destructive operation involving information loss.

\subsection{Entropy of the computing format}

Entropy is a measure of the amount of indeterminacy in information theory: the larger the information content
the lower the entropy and vice versa. Here, the computational entropy is defined as the normalized average of the
amount of indeterminacy of the arguments in an abstraction. 
A function assigns a single value to all arguments, hence it is a fully determinate object and its entropy is zero.
Partial functions have one or more unmarked columns, but the column index in such columns is 
fully determined too (i.e., 0), so the entropy of a partial function is also zero.

In the case of abstractions the degree of indeterminacy of an argument is inverse to its possible values.
Let $v_i$ be the number of values (i.e., the marked cells in column $i$), $x_i$ = $1/v_i$ 
and $n$ the number of arguments in abstraction $\Gamma$. In case the 
function is partial and $v_i = 0$ for column $i$, the corresponding 
$x_i = 1$. We define $e(\Gamma)$, the entropy of abstraction $\Gamma$, 
as follows:

\begin{mydef}
{\bf Entropy of an abstraction} \\
\vspace{0.4cm}
$e(\Gamma) = -1/n \sum_{i=1}^{n} log_2(x_i)$
\end{mydef}

The format of the Turing Machine is fully determinate
and it does not involve the entropy. However,
the notion becomes useful when an indeterminate format is incorporated.

\section{Associative Memory}
\label{associative-mem}

The extended distributed machine can be used to model an associative memory where contents are addressed through associations (e.g., \cite{Anderson1980}).
Artificial associative memories have been thought of as transfer functions 
were contents are represented as vectors, and have been designed and implemented 
as artificial neural networks trained with standard supervised machine learning methods  \cite{Hopfield, Hinton}. 
Associative memories have been also modeled within
the semantic networks paradigm (e.g., \cite{Quillian}) where concepts and their relations are represented as nodes and edges of a directed graph respectively, and associations are established through the paths between nodes. However, these graphs are represented through symbolic expressions within the standard symbol manipulation paradigm, and do not constitute distributed representations.

For the definition of the associative memory with the extended distributed machine the following considerations are taken into account:
\begin{enumerate}
\item The concepts or knowledge objects to be stored are represented by functions and abstractions 
(i.e, the objects of computations).\footnote{A discrete function of \emph{n} arguments can be thought of as a vector in a space of \emph{n} dimensions, and the representation
of contents in the present theory is analogous to the one used in associative memories modeled with neural networks.}
\item The external grid $G_{\psi}$ is used as an input and output buffer to place the objects to be registered, 
recognized or retrieved from the memory.
\item The control grid $G_{\phi}$ is used as the associative memory proper.
\item The units of form and content are functions and abstractions in the same format,
contents in the memory $G_{\phi}$ are accessed directly through contents in the grid $G_{\psi}$ and, consequently, the memory is associative. 
\end{enumerate}

The objects or scenes to be stored in the memory are characterized in terms of their salient features, such that each feature is characterized in turn
by a set of attributes with their corresponding values. Highly abstract feature-value structures appropriate for particular domains and modalities can be obtained through current deep learning techniques \cite{DeepLearning, Lecun, Lecun-nature}.
With this considerations we define a \emph{concrete image} of an object or scene
as the abstraction of the functions characterizing its salient features at a particular situation (e.g., in space and time). 
To capture a more stable representation we define an \emph{abstract image} as the abstraction of a set of concrete images. 
So, for instance, a visual scene can be captured by the abstraction of a number of concrete images produced within a short period of time, or through a number of variations of these objects due to observation points, illumination conditions, etc.

There is a supply of associative memory registers of arbitrary cardinalities only limited 
by the number of cells or basic processing units of the memory system as a whole, which is finite. So, a number of images can be stored in the memory.
We also consider that salient features of objects and scenes may depend on the modality through which a cognitive agent perceives such
object or scene, and particular associative memory registers may be associated to specific modalities of perception. Associative
memories can also represent modality independent concepts in terms of attribute-value pairs, although this is not addressed further in this paper.

The present associative memory offers the possibility of storing 
concrete and abstract images with a very large number of attributes that can be registered, 
recognized or recovered through associations. These operations are defined as follows:

\begin{mydef}
{\bf Memory Operations}
\end{mydef}

\begin{enumerate}
\item{Register:  Let $\Gamma_{\phi}$ be an image in $G_{\phi}$ and $\Gamma_{\psi}$ an image in $G_{\psi}$. Registering $\Gamma_{\psi}$ into  
$\Gamma_{\phi}$ is defined as (``$<=$" is an assignment operator):
\begin{enumerate}
\item{$\Gamma_{\phi} <= \lambda_{\phi}(\Gamma_{\phi},\Gamma_{\psi})$}
\item{$\Gamma_{\psi} <= B$}.
\end{enumerate}
}

\item{Recognition:  Let $\Gamma_{\phi}$ be an image in $G_{\phi}$ and $\Gamma_{\psi}$ an image in $G_{\psi}$. The image $\Gamma_{\psi}$ is in 
$\Gamma_{\phi}$ if the inclusion test of the reduction operation (clause 1 of Definition 2) holds.
}

\item{Retrieval:  Let $\Gamma_{\phi}$ be an image in $G_{\phi}$ and $\Gamma_{\psi}$ an image in $G_{\psi}$. Retrieving $\Gamma_{\psi}$ from  
$\Gamma_{\phi}$ is defined as:

\begin{enumerate}
\item {$\Gamma_{\psi}$ satisfies the inclusion test in relation to $\Gamma_{\phi}$}
\item {$\Gamma_{\phi} <= \lambda_{\phi}(\beta_{\phi}(\Gamma_{\phi}, \Gamma_{\psi})$}, $\Gamma_{\psi}$)
\item {$\Gamma_{\psi} <= \Gamma_{\phi}$}

\end{enumerate}
}
\end{enumerate}

Clause 1 of Definition 4 states that registering or inputing an image into associate memory is achieved by the abstraction operation directly. Recognition in clause 2 is achieved by testing that all the values for all arguments of the input or clue image are contained on the memory by material implication. Finally, retrieval is achieved by taking the clue out of the associative memory if the inclusion test is satisfied. However, memory retrieval involves also registering the input clue, which must be salient after a retrieval operation, and the final state of the memory is the abstraction of the reduced memory and the input clue. Finally, if the clue is contained in the memory the whole content of
the associative memory is placed into the output buffer, so contents in memory are selected by cue contents in the input buffer.
Such material may be used in other associative operations. The main intuition
underlying associative memories is that contents in memory are related to cue contents directly. 
These are \emph{the associations}. The use of functions as the basic units of form and content allow to model this intuition directly.

\subsection{A Case Study in Visual Memory}
\label{experiment}
\begin{figure}
\includegraphics[width=0.8\textwidth]{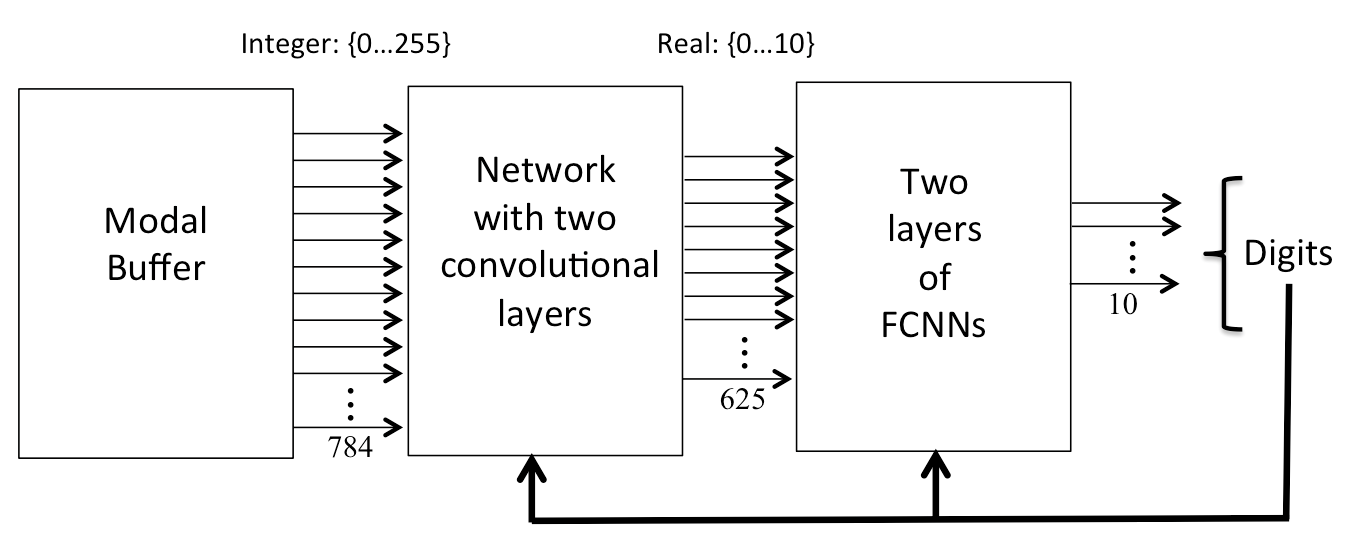}
\centering
\caption{Architecture for training the deep-learning network}
\label{training-architecture}
\end{figure}

The memory register and recognition operations were tested with a simulation experiment.\footnote{The data and source code for replicating the experiments, including the detailed results and the specifications of the hardware used, are available at \url{https://github.com/LA-Pineda/Associative-Memory-Experiments}}
The MNIST database of hand written digits from ``0'' to ``9'' was used.\footnote{http://yann.lecun.com/exdb/mnist/} 
Each digit is defined as a 28 x 28 pixel's array with 256 grey 
levels each and there were 70,000 instances available, where the 
ten digits are mostly balanced.
First, a deep neural network for classifying the digits was implemented.
The architecture follows closely LeCun's proposal \cite{Lecun} as illustrated in Figure \ref{training-architecture}.
The network has two convolutional layers mapping
the 784 (i.e., 28 x 28) inputs corresponding to the pixels to 625 outputs. Finally, two fully 
connected neural networks (FCNNs) mapping the 625 inputs to the 10 digits respectively were also defined.  
The network was trained with 60,000 pictures with a balanced number of digits. 

\begin{figure}
\includegraphics[width=1.0\textwidth]{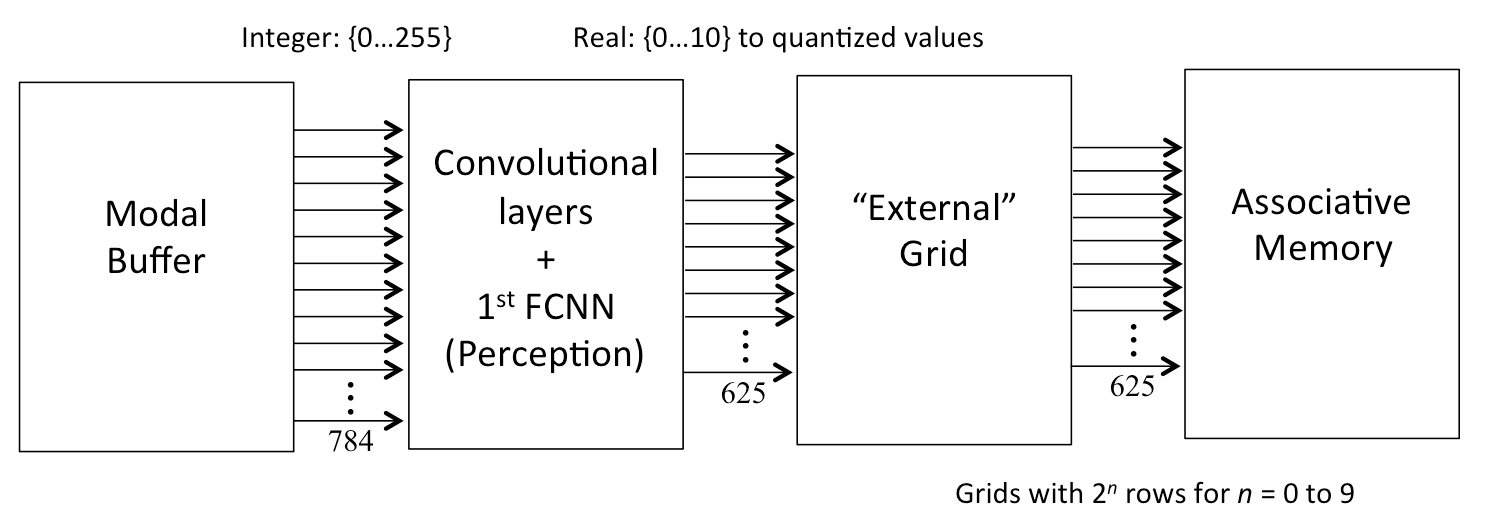}
\centering
\caption{Architecture using a deep-learning neural network and associative memory registers}
\label{associative-memory}
\end{figure}

For the experiments, the second fully connected neural network 
was substituted by the associative memory as illustrated in Figure \ref{associative-memory}.\footnote{The architecture is analogous to the basic
structure of the hybrid machine presented by Graves et al. \cite{Graves-Wayne, Graves-Wayne-nature}
although the vectors output by the deep neural network in such framework are read and written in
external RAM memory registers, that can be accessed by location and content,
instead of associative memory registers, as in the present proposal.}
Ten associative memory registers in a parallel array, one for each digit, were defined. 
Then, all training instances of the same digit were input through an abstraction operation into
the corresponding associative memory register. This input operation involved the two convolutional layers and the first FCNN. 
The outputs of the ``perceptual'' module were real numbers ranging between 0 
and 10 and the recognition experiments were run for memories with 
different quantized levels for the abstraction range, from 1 to 512, 
using all intermediate powers of 2.

\begin{figure} 
\includegraphics[width=0.8\textwidth]{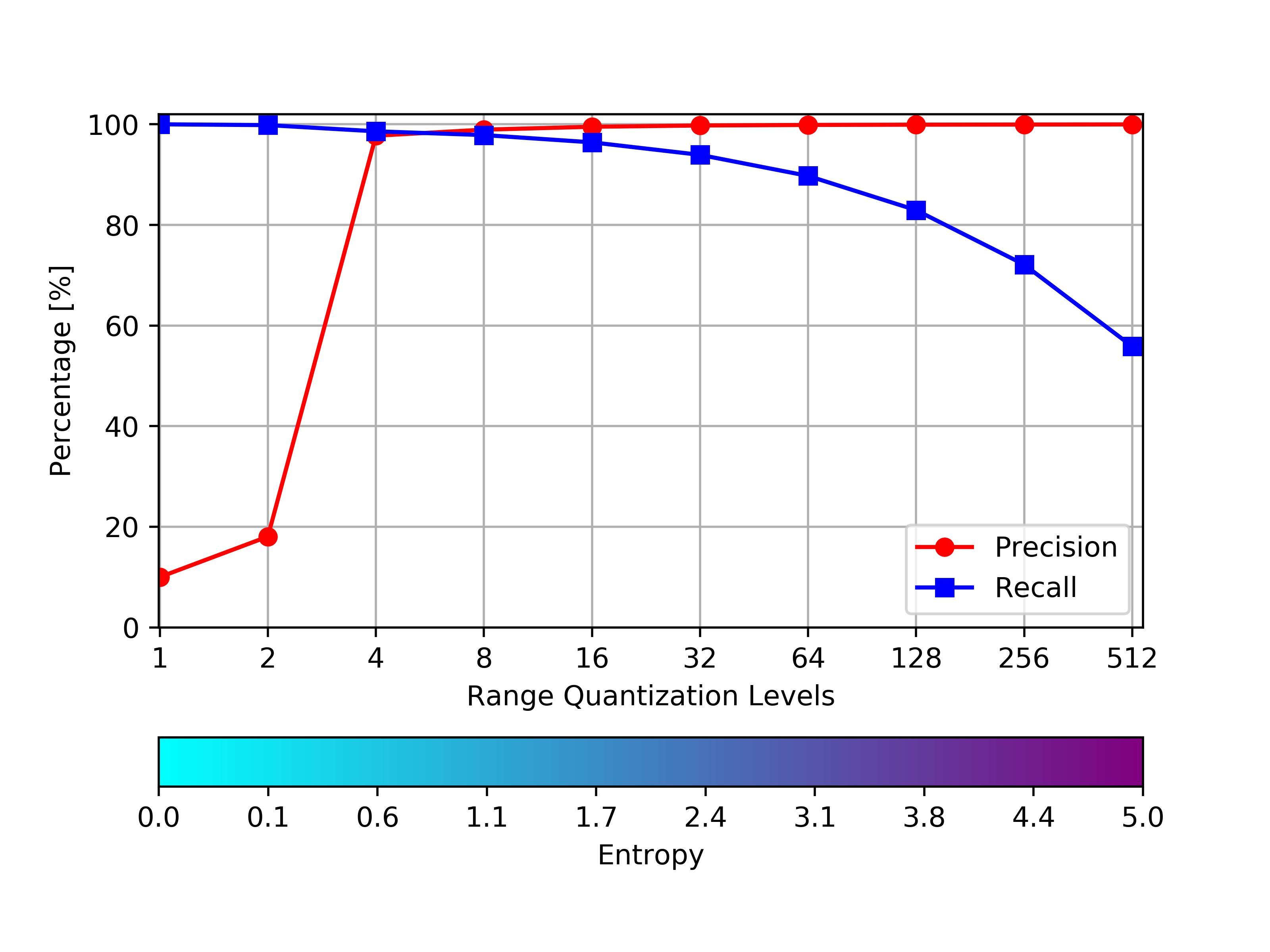}
\centering
\caption{Recognition experiment with the MNIST2 digit's database}
\label{experiment1}
\end{figure}

Once the memory was created the 10,000 test instances 
were submitted for recognition to the ten memories in parallel, using 
only the inclusion test of Definition 2.
The experiment was validated through an standard 10-fold cross-validation procedure. The averages of the precision, recall and entropy for the 10 grids 
are shown in Figure \ref{experiment1}.

As can be seen, precision grows very fast according to the 
grid's density, with a value very close to 1 (i.e., 100\%) with only four or 
eight quantization levels (0.977 and 0.989 respectively), and it 
is practically 1 from 32 levels onwards. Recall, on the other hand, 
is 1 for one quantized level (were all the information is confused) 
and declines with the grid's density. However, it is still very 
close to 1 with four or eight quantization levels (0.986 and 0.979 
respectively), with a very good compromise between recognition and 
recall. The confusion matrices also show that the rate of 
confusion for all digits is insignificant. In addition, the true 
negative identification for all digits was almost total. The graph 
shows that when the entropy is minimal everything is recalled, but 
precision is very low. However, the precision grows 
very rapidly with a small increment in the entropy, with a 
corresponding decreased in recall, but at a very reasonable rate. 
Finally, if the entropy is large, recall is decreased substantially. 
The experiment shows that an associative memory from 625 attributes to 
four to eight levels in the domain is quite effective for encoding the 
abstract image of hand written digits.

For this recognition task the fully-connected neural network produces similarly high 
recognition rates, and it is possible to argue that the associative memory 
is subsumed within the neural network or that memory is reduced to 
perception. However, the distinction is not about performance but 
about cognitive architecture; from this perspective it seems more natural 
to distinguish these two modules of cognition and to think of an associative 
memory not only as a filter or a classifier, but as a declarative storage 
of the concrete or abstract images. Furthermore,
recognition is achieved by computing the characteristic function of the clue
in relation to the abstraction in the memory.
So, classification is performed through abstraction and there is
no need of a classification algorithm.

\begin{figure} 
\includegraphics[width=0.8\textwidth]{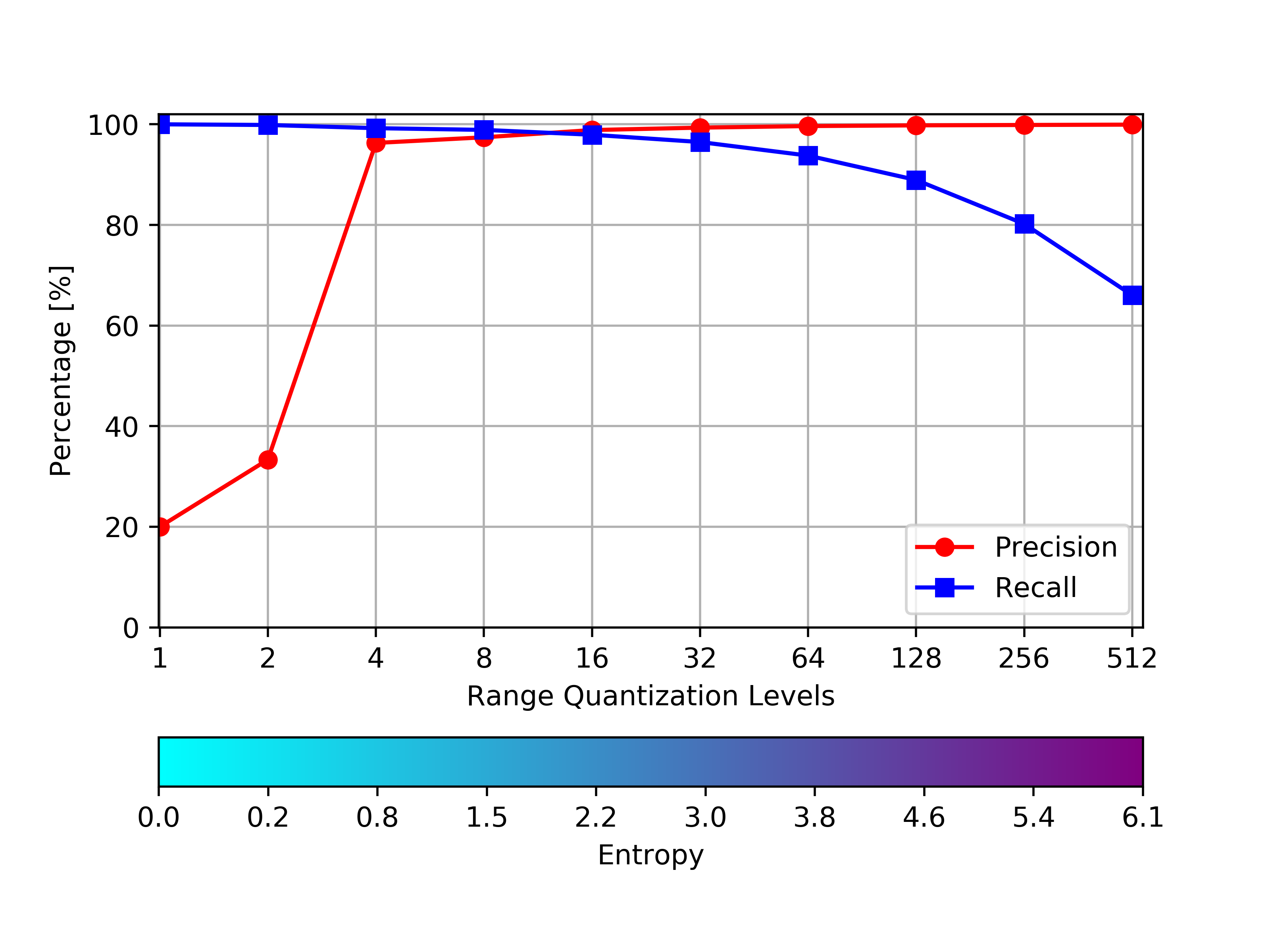}
\centering
\caption{Recognition experiment with two digits in the same memory}
\label{two_concepts}
\end{figure}

Another aspect of associative memory registers is that they can hold more than one abstract image at the same time,
such that basic cells or individual processing units contribute to the storage of more than one concept.
In this case the images are ``superposed" on the medium.
The average of the 10-fold cross-validation of the precision, recall and entropy of the 10 grids used for recognizing images with registers storing two digits
instead of only one (for the digit pairs ``0" and ``5", ``1" and ``6", ``2" and ``7", ``3" and ``8" and ``4" and ``9") 
are shown in Figure \ref{two_concepts}.
As before, the figures are satisfactory for only four quantized levels (0.963 and 0.992);
the precision is increased slightly with very good recall until eight levels (0.974 and 0.989) 
and the compromise is even better with 16 levels (0.989 and 0.979).
The entropy values for these 4, 8 and 16 levels are 0.822, 1.48 and 2.22 respectively.

However, augmenting the levels increases significantly the number of basic cells or individual processing units, 
with the corresponding increase in the cost of actual distributed architectures (as opposed to software simulations) or possibly of biological arrays. 
Also, the entropy is increased in relation to the associative memory registers that hold only one abstract image. This can be seen by
contrasting Figures \ref{experiment1} and \ref{two_concepts}. The present preliminary experiment suggests that the limit of abstract images that can be stored in an associative memory register will depend on a maximal entropy value 
above which the array is saturated and all the images are confused. Figure \ref{two_concepts} shows that the operational range is between 4 to 16 levels.
The choice of the granularity of an associative memory register will depend on a trade-off between the entropy 
at which the compromise between precision and recall is optimal and the total number of processing units of the array, which should be minimal due to their cost. 

The experiments show that information can be stored in a fully distributed or overlapped manner using a symbolic format --as opposed to connectionist systems; that information can be stored and retrieved locally through simple operations between adjacent cells using fully parallel processes, and that the represented content results from the interpretation of the overall array.

\section{Implications for the Theory of Computing}
\label{implications-for-computing}

The present extension of the Turing Machine has implications for the theory of computing and the way Church Thesis is understood. These are analyzed in relation to 1) the mode of computing that is employed by the TM, which is symbolic manipulation, and how other modes of computing can be characterized and distinguish from this basic sense; 2) a fundamental trade-off between distributed and local representations, which is presented and discussed, and 3) the notion of \emph{computing format} underlying the present theory.

\subsection{Modes of Computing and System Levels}
\label{modes-levels}

The symbolic manipulation mode of computing embedded in the Turing Machine and its difference from other modes of computing can be seen in terms of an adaptation of Newell's hierarchy of system levels \cite{Newell}. A system level is a level of abstraction at which a phenomenon can be studied --with its theoretical terms and laws of behavior-- independently of other system levels that analyze the phenomenon at different levels of granularity. Newell postulated that there is a system level, \emph{the knowledge level}, which stands at the top of this hierarchy. He claimed that the medium of this level is the knowledge itself and that the only rule of behavior at this level is the principle of rationality.
He also claimed that there is a system level directly below the knowledge level in which knowledge is represented formally as strings of symbols in digital computers, which is called the \emph{symbol level}. These representations are expressed through programming languages of all sorts and direct or cause the behavior of the computing machine. Strings of symbols are stored in memory registers and manipulated by processing units in a fully autonomous way, independently of human intervention. The symbol level in turn stands on top of all system levels involving the construction of digital computers, like the register transfer or computer architecture level, the logic circuit level, the electronic circuit level, the device level and the physical level --involving the physical phenomena-- at the bottom.

Newell claimed that the knowledge level emerges from the symbol level in digital computers, but here we depart from such claim --with its implications-- and only pose that the knowledge level consists on the interpretations made by people of the symbolic structures held by computers. So, the knowledge level in the present sense is human knowledge that can be expressed through diverse means, like natural language, diagrams and informal descriptions of all sorts. In this view the knowledge level involves the attribution of meanings to symbolic structures, including the interpretation conventions of standard input and output configurations, but the symbol level is fully mechanical and does not involves interpretations or meanings. The knowledge level, including emergent entities and their properties, cannot be reduced to the symbol level, but this latter level is compositional and can be reduced to all lower system levels down to the physical level, in accordance with Newell's proposal.

The symbol level characterizes the  \emph{mode of computing} of the Turing Machine which is \emph{symbolic manipulation} as opposed to alternative modes that use specific physical properties to compute continuous functions, like analogical computers that model dynamical systems. Optical and quantum computers may be considered as other modes of computing in this latter sense too. Even sensors and transducers of all sorts can be thought of as devices that compute particular functions on the basis of diverse physical phenomena, each with a particular mode of computing. These machines involve a reduced number of systems levels: the knowledge level at which the input and output signals are interpreted by people, the device level, which corresponds to the particular mode of computing, and the physical level which involves the physical phenomena used by the particular device.

In general, computing machines have a system level directly below the knowledge level that characterizes its particular mode of computing. 
Such system level is called here \emph{the functional or computational system level}. 
The symbol level is the functional or computational level of digital computers or TMs. Non-symbolic modes of computing have a system level directly below the knowledge level which is non-symbolic.

Algorithms are symbolic procedures for mapping arguments into values effectively, so this notion is intrinsically tied up with the symbol level and symbolic computation. Although sometimes computations made through non-symbolic modes are said to be algorithmic, this use is vague and informal. For instance, there is no well-defined sense in which an analogical computer can be said to compute an algorithm. Physical phenomena are not computing machines by themselves, and a device or a physical phenomenon becomes a computing engine only when its inputs and outputs are interpreted as arguments and values of a function by an external interpreter capable to make such semantic attribution.

According to the present discussion there is an open ended number of modes of computing, that can be exhibited by natural organisms or engineered artifacts. The Turing Machine is one of such machines of one particular mode which is symbolic manipulation. The present distributed extension of the TM augments the symbolic manipulation mode with indeterminate formats and the computational entropy.

\subsection{The trade-off between Distributed and Turing Machine Computations}
\label{trade-off}

In Shannon's information theory the entropy
is the expected value of the amount of information in a message in a communication system.
As the length of a message is the inverse of its information
content, Shannon's entropy is defined operationally as the average
size of the length of a message. In the present theory
the amount of indeterminacy of an argument
is proportional to its possible values --in excess of one-- within an abstraction, so this parameter
is analogous to the length of a message. Definition 3
computes the normalized average of this parameter,
and the computational entropy presented in this paper 
is analogous to Shannon's entropy.

In communication theory the entropy is useful to determine the capacity of the
channel so all messages can go through in the expected time. If the entropy were not considered
the capacity of the channel would have to be large enough to
pass all messages, regardless their probability, which may be very low,
with the corresponding penalty in the cost of the system as a whole.
Likewise, not considering the computational entropy
is playing safe, as there is no information loss, but 
security always comes with a high price.

This suggests that there is a fundamental trade-off between distributed 
and Turing Machine computations. The former capture relations at the level
of the format and computations are local to the basic units of memory or processing; for these reasons
distributed representations are highly expressive --can express large amounts of relational and qualitative information, 
including complex individual concepts represented through abstractions-- and can be computed very effectively; however
they are indeterminate --entropic-- and have to face information loss. 
The standard Turing Machine, on the other hand, is information preserving so computations
are precise, but has to assume the compromise between expressivity --for capturing large abstractions in language-- 
and effective computation, the costs associated to the complexity of algorithms
and the knowledge representation trade-off \cite{Levesque-brachman, Levesque}.

The trade-off between distributed and standard Turing Machine computations
can be seen in relation to the entropy. The Turing Machine is simply
the case of fully determinate computations and its entropy is zero. 
However, it is possible to conceive distributed formats that allow different degrees
of indeterminacy such that there is an entropy level in which
the compromise between expressivity and tractability is satisfactory.
This may be the case for natural computations, that are
massively distributed, use a very expressive format and are carried on very fast, often in real time,
and nevertheless are highly precise.

\subsection{The notion of computing format}
\label{format}

The present trade-off helps also to clarify the notion of \emph{computing format}: it is perhaps possible to see an indeterminate format as a determinate one that is inspected and explored through a non-deterministic process. This strategy would require to explore a number of paths, possibly large, and to keep a number of copies of each state of the associative memory to prevent information loss. However, the compromise between expressivity and computational costs of determinate formats would have to be assumed, and hence the trade-off holds. A computing format involves the shape of the medium, the alphabet, the scanning protocol and the interpretation conventions --including the notation-- and changing these conventions, even if the rest remains the same, changes the computing format, and the place of the machine in the trade-off.

\subsection{Implications for Church Thesis}
\label{implications-for-Church-Thesis}

Church Thesis states that the Turing Machine computes the full set of computable functions, which corresponds to the set of functions that can be computed intuitively by people, and that any other general theory or machine computes the same set of functions and is equivalent to the TM for this reason. The thesis was motivated by the need to formalize the notion of effective calculability which was a mathematical concerned at the time --i.e., the calculations that can be performed by people given the appropriate physical and time resources. Church Thesis is often extrapolated into the statement that the Turing Machine is the most powerful model of computing in any possible sense. However, this extension is problematic because although it stands solid in relation to \emph{what} computations can be made (i.e., the set of computable functions and the impossibility of computing uncomputable functions, like the halting problem) it does not in relation to \emph{how} computations are performed.

In this latter sense different machines --that may be equivalent in the set of functions that they can compute, even if this is the full set of computable functions-- assume different trade-offs, and can be more powerful than others in a given particular sense. Indeed, computing machines and representation schemes are singled out by the particular trade-offs that they assume. Algorithms depend on the computational format, and the difference between alternative theories of computable functions, like recursive functions theory, the $\lambda$-calculus, extended standard TMs with several tapes or grids, the Von Neumann architecture, etc., depends on the format that they pose and the focus of the theory. In particular, the trade-off between distributed and standard TM computations illustrates the strengths and limitations of the corresponding machines, despite that both can compute the full set of computable functions. So, there is a particular sense in which a machine can be more powerful than the standard TM due to its place in the trade-off.

It is also unclear whether the strong version of Church Thesis applies to machines whose mode of computing is different from symbolic manipulation. Such devices compute specific functions or limited classes of functions, possibly continuous, taken advantage of natural phenomena directly, and can be more powerful than practical implementations of digital computers in a particular sense, like analogical or quantum computers, that make computations almost instantly. Of course these computations can also be made by digital computers through an appropriate discrete approximation and conform to Church Thesis in this respect, but comparing algorithms with physical phenomena and abstract devices with concrete ones are category mistakes. So the strong version of Church Thesis is problematic in this sense too.

In a yet another sense, if the brain turns out to be a machine using a number of specialized continuous devices --whose mode of computing is non-symbolic-- for the set of functions that is able to compute, that underlie all mental processes, and if the TM fails in the end to model the same processes at an adequate level of performance, Church Thesis would still hold in relation of the set of functions that can be computed, but there would be a machine that is more powerful than the TM in a fundamental way.

The distributed extension of the Turing Machine presented in this paper does not augment the set of functions that can be computed effectively, and conforms to Church Thesis, but it does extend the symbolic manipulation mode of computing, supports distributed representations at the symbol level directly and uses explicitly the trade-off between distributed and standard TM computations.

\section{Implications for Artificial Intelligence and Cognition}
\label{implications-for-cognition}

Indeterminate formats, the computational entropy and the distributed extension of the TM have implications for diverse aspects of Artificial Intelligence and Cognition; some of these are briefly discussed next.

\subsection{Symbolic versus Sub-Symbolic Information}
\label{sym-versus-subsym}

The standard Turing Machine has been used in Artificial Intelligence and Knowledge Representation to express both ``symbolic" and ``sub-symbolic" information.
The former term is used when the information is expressed through declarative languages with a well-regimented interpretation, like logically oriented knowledge-bases or semantic networks, and even logical and functional programming languages like Prolog and Lisp, while the latter is used when the information is expressed through algorithms or data-structures whose internal structure is opaque, like ANNs, procedural representations, genes in genetic algorithms or clusters in vectors' spaces. In this latter sense symbolic and sub-symbolic systems are often said to be ``representational" and ``non-representational" respectively.\footnote{Although this terminology is informal and somehow contradictory; for instance, distributed and procedural representations are ``non-representational".} However, highly expressive declarative languages underlie a virtual machine, which in turn can be implemented in a more basic virtual machine, down to the actual physical machine that implements the computation. So, information that at certain level is taken to be symbolic can be seen as non-symbolic at another lower or virtual machine, but the reduction is well-defined and as the mode of computing is symbolic manipulation all the information is in the end symbolic.

The distinction can be seen in terms of the \emph{knowledge representation hypothesis} \cite{Brian-Smith}. According to this, a scheme is representational if it is possible for an external observer --a human interpreter-- to give a propositional account of the knowledge that the process exhibits --to render such knowledge as a set of propositions-- but independently of such semantic attribution, the knowledge is formal and causal of the behavior engendered by such process. So, reasoning and theorem-proving are symbolic and representational, but learning and classifying objects through ANNs are sub-symbolic and non-representational. 

In this AI sense, the functions and abstractions in the control state and the external grid of the distributed extension of the TM, and the attribute-value structures in the associative memory registers, can be considered sub-symbolic information.

\subsection{Connectionism versus Symbolic Cognition}

Digital computers are used for developing and testing models of Cognition. The availability of Turing's model and the Von Neumann Architecture inspired the symbolic model of cognition, that has a strong syntactic and semantic orientation. However, this approach was questioned by Connectionism and ANNs on the grounds that models of perception, thought and motor behavior requiere non-symbolic distributed representations, as mentioned in the introduction of this paper. Such criticisms have been contested by the symbolic approach due to the limitations of connectionist systems for representing and operating on symbolic structures and for storing and retrieving such structures in memory \cite{Fodor-Pylyshyn} and although in principle ANNs can capture syntactic structure (e.g. \cite{Chalmers,Lecun-nature}) such kind of proposals are still very limited. 

As stressed by Fodor and Pylyshyn \cite{Fodor-Pylyshyn} both the symbolic and the connectionist paradigms are \emph{representational} in the sense that the formal structures or units of form written up on the tape or stored in memory are interpreted as units of content. The difference is that in the former the semantic attribution is made upon linguistic expressions while in the latter it is made upon the distributed set of nodes or neurons.

The relation between symbolic and connectionist systems has been subject of much debate in cognitive psychology and philosophy of mind in terms of system levels (e.g., \cite{Rueckl}). In particular Marr proposed that an adequate explanation of a psychological process involves three levels: the computational, the algorithmic and the implementational \cite{Marr}. The symbolist side sustains that the computational level is characterized by a symbol-level representation with a compositional syntax and semantics \cite{Fodor-Pylyshyn}, and particular models focus on the algorithms that manipulate such symbolic structures. Supporters of connectionism argue on their part that explicit symbolic structures are not needed and that this paradigm provides an extended class of algorithms that reflects better the structure of natural neural networks. But the symbolist side argues back that connectionism only provides an alternative implementation \cite{Fodor-Pylyshyn} --despite that whether there is a reduction from symbol systems to ANNs is still an open question-- and there seems to be an impasse between these opposite positions. 

However, ANNs are normally simulated in TMs and although the distributed property is described at the knowledge level through textual descriptions, formulas and diagrams with their labels, the actual specification at the symbol level is local. Hence, the distributed property itself is only simulated. From this perspective ANNs can be thought of as virtual machines that use the TM as their actual physical machine; consequently, ANNs have to be placed with the standard Turing Machine in the entropy free side of the trade-off between distributed and TM computations. To overcome this limitation, a functional or computational system level for Connectionism --using a non-symbolic mode of computing-- that does not rely on the TM should be provided.

The connectionist side could argue at this point that ANNs model the information processes of the brain directly, but when asked what kind of computational machine is the brain and what is its mode of computing, the answer would have to be that it is the one described by the ANNs, incurring in circularity. The Turing Machine, on its part, is a general theory of computable functions that provides a physical machine, that computes functions through symbolic manipulation.

There is also the claim that representations are not needed and that intelligence can be modeled through algorithms organized in sub-summed architectures \cite{Brooks}. This non-representational view underlies Embodied Cognition \cite{Anderson-b:2003} and Enactivism \cite{Froese} too. These paradigms emphasize the role of the body and the interaction with the environment in learning, experience and consciousness, and hold that the underlying information processes are not symbolic. Hence the rejection of the Turing Machine. However, these approaches lack an explicit model of computing and do not specify what is the mode of computing of the brain or the body. Despite this, they do use ANNs and other classes of algorithms as standard modeling tools, but as no semantic attribution is made neither to data-structures nor to the operations performed upon them, the computational structures and processes play an implementational role only.

The present distributed extension of the TM, on its part, do supports the distributed property and distributed models can be implemented directly in an actual machine.  
The superposition of the abstraction operation in the distributed extended machine is an instance of an actual non-compositional operation. The associative memory presented in this paper illustrates a distributed representation using an indeterminate and entropic format --that can be computed by a fully parallel architecture through direct operations between simple processing units. The machine operations change the functions and abstractions that are computed and the machine is interactive for this reason, attending better the concerns of Embodied Cognition and Enactivism, although it is neutral to whether the units of form are assigned an interpretation in cognitive modeling.

More generally, the symbolic mode of computing manipulates units of form mechanically but the semantic attribution is made by people. Whether there is a natural mode of computing capable of making such attribution underly genuine representations, feelings, emotions and consciousness. Current computing machines do not have the structural and/or functional elements that explain such phenomena. If such mode of computing were ever discovered the knowledge and computational system levels would be identified, and such machine would be more powerful than the Turing Machine too.

\subsection{Cognitive Architecture}

The experiment in Section \ref{experiment} suggests a cognitive architecture in which perception
is modeled through deep neural networks but concepts or images are stored and retrieved in associative memory registers.
The input and output to perception are attributes and values
that are best thought of as sub-symbolic information in the AI sense, but the content of an associative memory register as a whole is already symbolic and can be labeled with a symbol. In addition, images of different modalities, like the visual and the acoustic one, can be given the same linguistic label, such that the symbol's content is multimodal. 
For this an associative memory can be thought of as a coin with to sides, one presenting the information of the world to perception and motor behavior sub-symbolically, and the other presenting the same information to thought and language as fully articulated symbols. Such architecture is consistent with Kosslyn's proposal \cite{Kosslyn} in which an associative memory coupled with modality specific imagery buffer plays a central role.

\subsection{Imagery}

Traditional models of cognition hold that representations that are built in the mind have a propositional or linguistic character \cite{Pylyshyn}; however,
there has been claimed that in addition to propositions there are mental images \cite{Kosslyn, Shepard-Cooper} giving raise to the imagery debate \cite{Tye}. 
The propositional side argues that imagery confuses the concrete modalities of perception and/or motor behavior with the cognitive level, which needs to be fully abstract, and that propositions reflect the actual computing format. According to this if the linguistic format of the TM is the only one that there is, the propositional camp is right. However, if this is one among several possible computing formats, which may be underdetermined and entropic, imagery is plausible. Images would reflect the actual format carried on with the computation. In this view the functional mental representation would be identified with the output of the perceptual stage but not with its input, satisfying the concerns of the propositional side \cite{Pylyshyn}.

\subsection{Propositional versus Analogical Representations in AI}

An associative memory in conjunction with images also informs the old Artificial Intelligence (AI) debate between
propositional and analogical representations \cite{Sloman, Hayes}. The propositional side argued that knowledge expressed
through schemes different than language, like maps or diagrams, can be reduced to propositions.
Here, the argument was that the medium, like the space, can also be represented in a linear format, which is the format of
the Turing Machine. Methodologies and case studies for exploring this position and its limits can be seen in \cite{PinedaCL, PinedaAIJ}. 
However, if the linguistic format
is one among a range of possible computing formats such medium reduction is not necessary, not useful and perhaps not possible.

\subsection{The Knowledge Representations Trade-Off Paradox}

The introduction of an associative memory also helps to dissolve a paradox related to the
so-called knowledge representation trade-off \cite{Levesque-brachman, Levesque}.
This trade-off states that there is a fundamental compromise between
the expressive power and the tractability of a representational language: the larger the expressive power the
less tractable. This is also a consequence of a more fundamental computational
property embedded in Chomsky Hierarchy  \cite{Hopcroft} such that the larger the expressive power of a formal language
the more expensive to compute whether a string is within its extension. 
More generally, as the expression of abstractions requires a large expressive power, computing with abstractions is hard,
and sometimes impossible. 

However, this conflicts with the intuition that the solution of hard problems requires abstraction. 
If a problem is easy it can be solved through concrete thinking,
but if it is hard, solving it requires abstract thinking. Abstractions may be hard to build or learn but
once they are in place, thinking with them is easy. Expert performance relies on abstractions.
The paradox comes from the fact that reasoning with abstractions is not only easier for expert thinkers but also essential for the solution of hard problems, while the knowledge representation trade-off states that expressing higher abstractions makes computations harder if not impossible \cite{PinedaAIJ}.

Chomsky's Hierarchy and the knowledge representation trade-off assume that knowledge is propositional
and hence expressed through language, but the inclusion of an alternative computing format, associative memory and abstract images
offer an alternative to abstract thinking. For instance, if the premises and the conclusion of an argument were represented as abstractions in the associative memory, verifying that the argument follows could be achieved through a direct memory operation. From this perspective the paradox may be dissolved.

\subsection{Impact on Learning and Creativity}

The inclusion of an associative memory and imagery has also consequences for learning and creativity. 
Information enrichment is implicit to the abstraction operation and is responsible of the generalization provided by the present scheme, in line with other kinds of distributed representations that support some or another form of generalization. 
However, information enrichment depends on a property of the computing format in opposition to standard machine learning algorithmic
techniques that rely on similarity measures, either implicit or explicit, between the objects to be learned or classified. 
Also, the combination of abstraction
and reduction modify the abstract images stored in the associative memory registers,
and the gain and loss of information can impact in the generation of new concepts. The operations
within the associative memory registers are not transparent and their products could be presented as ready made objects to
other layers of representation and thought.

\subsection{Natural Versus Artificial Computations}
The proposal that all mental processes can be model computationally, 
which was originally made explicit by Turing \cite{Turing}, and has 
inspired most work in AI and related disciplines, presupposes the 
intensive development and use of algorithms. The notion of algorithm 
is commonly associated to arithmetic, algebraic or more general 
mathematical methods, mostly for making calculations or resolving 
equations that people cannot solve naturally, without the support 
of an external representation, like paper and pencil, possibly in 
large quantities, and a considerable amount of time. The structure 
and basic set of actions of the Turing Machine reflect this intuition 
very directly. Mathematical models require algorithms, and in this 
setting, algorithms seem to be the only way to make computations in 
practice. The invention of computing machines permitted to take this 
intuition to its ultimate consequence, and computers amplify people 
with the power of making extreme calculations that would be impossible 
to do without mechanical artifacts, with great benefits for science and 
technology and their effects in all ambits of society, as we see in 
the world today. 

However, algorithmic computations in the 
format of the Turing Machine are artificial or engineered as opposed 
to computations that can be performed by people and living organisms with
a developed enough neural system, which 
we refer here as ``natural computations". AI and related disciplines have 
focused in modeling the latter through the tools and methodology of 
the former. Even artificial neural networks simulated in the Turing Machines
use very costly algorithms.

Nevertheless, although there have been impressive advances in 
algorithmic models of natural faculties --that humans exercise 
without an apparent effort, like vision, language and motor behavior-- 
there is still a huge gap between what higher developed animals and machines 
can do, and it is not even clear that these faculties can be modeled 
to the human standards with pure algorithmic force. For this, the availability of an associative memory 
with direct accessibility and transparent 
interpretation may be essential to realistic models of natural 
computing.

Scientific theories aim to be fully adequate models of the phenomenon or phenomena that they mean to capture. 
In the case of computing, a very strong current of opinion sustains that the Turing Machine provides a fully adequate model 
of computing. Whether this is the case depends on the underlying notion of computing that one has in mind.
If computing is thought of as a game, like chess, and the rules of the game are defined by the Turing Machine,
the theory of TMs is a fully adequate model of computing in the same way the rules of chess provide
an adequate model for chess. But computing is more often thought of as a human invention and the TM as a fully adequate abstract model of all machines 
that use information preserving determinate formats, that have and will be ever designed. This is the intended sense of Church Thesis.
However, computing can also be thought of as the information processes performed by arbitrary distributed systems, including
the brains of living entities with a developed enough neural
system and, consequently, as a natural phenomenon or phenomena whose study becomes a natural science.
Humans and other animals are very bad at computing algorithms unless they are very simple,
and the algorithmic power comes from machines; but at the same time, people and other animals
can make very complex natural computations with a level of competence that is largely beyond the capabilities 
of the most sophisticated current machines.
Hence, the strategies of natural computing may not rely in complex algorithms.
The present theory and discussion suggest
to allow indeterminate formats and to include the computational entropy 
towards the formulation of a comprehensive theory that encompasses natural and artificial computing.

\section{Acknowledgments}
The author thanks Gibr\'an Fuentes for useful comments and the design of the experiment in Section \ref{experiment}, to Ra\'ul Peralta for its implementation and to Iv\'an Torres for its verification.

%\nolinenumbers

%\section{Bibliography styles}
%There are various bibliography styles available. You can select the 
%style of your choice in the preamble of this document. These styles are 
%Elsevier styles based on standard styles like Harvard and Vancouver. 
%Please use Bib\TeX\ to generate your bibliography and include DOIs 
%whenever available.
%Here are two sample references: \cite{Feynman1963118,Dirac1953888}.

%\section{References}

%\bibliography{mybibfile}

\begin{thebibliography}{35}

\bibitem{Abramsky}
Abramsky, S., A Structural Approach to Reversible Computation, Theoretical Computer Science 347(3):441-464, 2005. doi:10.1016/j.tcs.2005.07.002

\bibitem{Anderson1980}
Anderson, J. R., Bower, G. H., Human Associative Memory: A Brief Edition, Lawrence Erlbaum Associates, Publishers, Hillsdale, New Jersey, 1980.

\bibitem{Anderson:2003}
Anderson, M., McCartney, R., Diagram processing: Computing with diagrams, Artificial Intelligence 145:181--226, 2003.

\bibitem{Anderson-b:2003}
Anderson, M. L., Embodied Cognition: A field guide, Artificial Intelligence 149:91-130, 2003.

\bibitem{Anderson:2010}
Anderson, M., Furnas, G., Relating Two Image-Based Diagrammatic Reasoning Architectures.
 in A. K. Goel, M. Jamnik, Narayanan, N. H. (Eds.), Diagrammatic Representation and Inference, 
Proceedings of the 6th International Conference, Diagrams 2010, LNCS  6170, Springer, pp. 128-143, 2010.

%\bibitem{Hyperproof}
%Barwise, J., Etchemendy, J., Hyperproof, CSLI Lecture Notes, vol. 42, CSLI Publications, Stanford, CA, 1994.

\bibitem{DeepLearning}
Bengio, Y., Learning Deep Architectures for AI, Foundations and Trends in Machine Learning 2(1):1--127, 2009. doi:10.1561/2200000006.

\bibitem{Bennet}
Bennet, C. H., Logical Reversibility of Computation, IBM Journal of Research and Development 17:525--532 (November, 1973).

\bibitem{Brooks}
Brooks, R., Intelligence without representation. Artificial Intelligence (47):139--159, 1991.

\bibitem{Boolos-Jeffrey}
Boolos, G. S., Jeffrey, R. C., Computability and Logic, Third Edition, Cambridge University Press, Cambridge, 1989.

\bibitem{Chalmers}
Chalmers, D. J., Syntactic Transformations on Distributed Representations, Connection Science 2(1--2):53--62, 1990. doi:10.1080/09540099008915662

%\bibitem{chandra}
%Chandrasekaran, B. Diagrammatic Representation and Reasoning: Some Distinctions. Invited paper, AAAI Fall 97 Symposium Series, Diagrammatic Reasoning, Boston, MA, 1997.

%\bibitem{Dreyfus}
%Dreyfus, H. L., From Micro-Worlds to Knowledge Representation: AI at an Impasse, in: R. Brachman, H. Levesque (Eds.), Readings in Knowledge Representation, Morgan and Kaufmann, Los Altos, CA, 1985, pp. X-Y.

\bibitem{Funt}
Funt, B. V., Problem solving with diagrammatic representations. Artificial Intelligence 1980. 13: 201--230.

%\bibitem{Johndon-Laird}
%Johnson-Laird, P. N., Mental Models: Towards a Cognitive Science of Language, Inference, and Consciousness, Cambridge University Press, Cambridge, 1983.

\bibitem{Froese}
Froese, T., and Ziemke, T., Enactive artificial intelligence: Investigating the systemic organization of life and mind. Artificial Intelligence,
173:466--500, 2009.

\bibitem{Fodor-Pylyshyn}
Fodor, J. A., Pylyshyn, Z. W., Connectionism and cognitive architecture: A critical analysis, Cognition 28(1-2):3-71, 1988.

\bibitem{Graves-Wayne}
Graves, A., Wayne, G., Danihelka, I., Neural Turing Machines, arXiv:1410.5401v2 [cs.NE], 2014.

\bibitem{Graves-Wayne-nature}
Graves, A., Wayne, G., et al., Hybrid computing using a neural network with dynamic external memory, Nature 538:471--476 (27 October 2016). doi:10.1938/nature20101.

\bibitem{Hayes}
Hayes, J. P., Some problems and non-problems in representation theory, in: R. Brachman, H. Levesque (Eds.), Readings in Knowledge Representation, Morgan and Kaufmann, Los Altos, CA, 1985, pp. 3--22.

\bibitem{Hinton}
Hinton, E. H., Anderson, J. A., Parallel Models of Associative Memory, Psychology Press, New York, 1989.

\bibitem{Hinton-1986}
Hinton, G. E., McClelland, J. L., Rumelhart, D. E., Distributed Representations (Chapter 3), in: Rumelhart and McClelland (Eds.) Parallel Distributed Processing, Explorations in the Microstructure of Cognition, Vol.1: Foundations, The MIT Press, Cambridge, Massachusetts, 1986, pp. 77--109.

\bibitem{Hopcroft}
Hopcroft, E., Motwani, R., Ullman, J., Introduction to Automata Theory, Languages and Computation (Second Edition) John Addison Wesley, 2001.

\bibitem{Hopfield}
Hopfield, J. J., Neural networks and physical systems with emergent collective computational abilities, Proceedings of the National Academy of Sciences of the USA 79(8):2554--2558, April 1982.

%\bibitem{Hyotyniemi} Hyotyniemi, H., Turing Machines are Recurrent Neural Networks, Proceedings of STeP'96, Jarmo Alander, Timo Honkela and Matti Jakobsson (eds.), Publications of the Finnish Artificial Intelligence Society, pp. 13-24, 1996.

\bibitem{Kosko}
Kosko, B., Bidirectional Associative Memories, IEEE Systems, Man, and Cybernetics 18(1):49--60, 1980.

\bibitem{Kosslyn}
Kosslyn, S. M., Thompson, W. L., Ganis, G., The Case for Mental Imagery, Oxford Univ. Press 2006.

%\bibitem{Lai-chess}
%Lai, M., Giraffe: Using Deep Reinforcement Learning to Play Chess, Imperial College London, Master's dissertation: arXiv:1509.01549v2 [cs.AI], 2015.

%\bibitem{Simon}
%Larkin, J. H., Simon, H., Why a diagram is (sometimes) worth ten thousand words, Cognitive Science 11 (1987) 65--99.

\bibitem{Lecun}
LeCun, Y., Bottou, L., Bengio, Y.,  Haffner, P., Gradient-based learning applied to document recognition, Proceedings of the IEEE 86(11):2278--2324, 1998.

\bibitem{Lecun-nature}
LeCun, Y., Bengio, Y., Hinton, G., Deep Learning, Nature 521:436--444, 2015 (27 May 2015). doi:10.1038/nature14539.

\bibitem{Levesque}
Levesque, H. L., Logic and the complexity of reasoning, Journal of Philosophical Logic 17:355--389, 1988.

\bibitem{Levesque-brachman}
Levesque, H. L., Brachman, R., A fundamental tradeoff in knowledge representation and reasoning, in: R. Brachman, H. Levesque (Eds.), Readings in Knowledge Representation, Morgan and Kaufmann, Los Altos, CA, pp. 41--70, 1985.

\bibitem{Leeuwen-Wiedemann-2000}
van Leeuwen, J., Wiedemann, J., On algorithms and interaction. In: M. Nielsen and B. Rovan (eds.) Mathematical Foundations of Computer Science 2000, 25th Int. Symposium (MFCS 2000), Lecture Notes in Computer Science, vol. 1893. Springer, Berlin, pp. 99–113, 2000.

\bibitem{Leeuwen-Wiedemann-2001}
van Leeuwen J., Wiedermann J. The Turing Machine Paradigm in Contemporary Computing. In: Engquist B., Schmid W. (eds) Mathematics Unlimited — 2001 and Beyond. Springer, Berlin, Heidelberg

%\bibitem{SIFT}
% Lowe, D., Distinctive image features from scale-invariant keypoints. International Journal of Computer Vision 60(2): 91--110, 2004.

\bibitem{Marr}
Marr, D. Vision: A Computational Investigation into the Human Representation and Processing of Visual Information. New York: Freeman. (1982).

\bibitem{Newell}
Newell, A. The Knowledge Level, AI Magazine, 1981. 

\bibitem{PinedaCL}
Pineda, L. A., Garza, G., A Model for Multimodal Reference Resolution, Computational Linguistics 26(2):136--192. MIT Press, 2000.

\bibitem{PinedaAIJ}
Pineda, L. A., Conservation Principles and Action Schemes in the Synthesis of Geometric Concepts, Artificial Intelligence 171(4):197--238, 2007.

%\bibitem{Patente}
%Pineda, L. A., Sistema de memoria asociativa basada en la abstracci\'on diagram\'atica de contenidos especificados como estructuras atributo-valor. Patent Application MX/a/2015/005878, Instituto Mexicano de la Propiedad Industrial (IMPI), 2015.

\bibitem{Pylyshyn}
Pylyshyn, Z. W., What the mind's eye tells the mind's brain: A critique of mental imagery, Psychological Bulletin 80, 1–24, 1973.

\bibitem{Quillian}
Quillian, M. (1968). Semantic Memory, in M. Minsky (ed.), Semantic Information Processing, pp. 227--270, MIT Press, 1968; reprinted in Collins \& Smith (eds.), Readings in Cognitive Science, section 2.1.

\bibitem{Rueckl}
Rueckl J.G., Connectionism and the Notion of Levels. In: Horgan T., Tienson J. (eds) Connectionism and the Philosophy of Mind. Studies in Cognitive Systems, vol 9. Springer, Dordrecht, 1991.

\bibitem{Rumelhart}
Rumelhart, D. E., McClelland, J. L. and the PDF Research Group. Parallel Distributed Processing, Explorations in the Microstructure of Cognition, Vol.1:  Foundations, The MIT Press, Cambridge, Massachusetts, 1986.

\bibitem{Shepard-Cooper}
Shepard, R., Cooper, L., Mental images and their transformations, Cambridge, MA: MIT Press, 1982.

\bibitem{Shimojima} Shimojima, A., A Logical Analysis of Graphical Consistency Proofs, in Logical and Computational Aspects of Model Based Reasoning, L. Magnani and N. Nersessian (eds), Applied Logic Series 25, Dov Gabbay (Ed.), Kluwer-Academic, Dordrecht, Holanda, pp. 93--109, 2002.

\bibitem{Siegelmann}
Siegelmann, H. T., Sontag, E. D. On the Computational Power of Neural Nets, Journal of Computer and System Sciences 50(1):132--150, 1995. doi.org/10.1006/jcss.1995.1013

\bibitem{Siegelmann97analogcomputation}
Siegelmann, H. T.,  Fishman, S. Analog Computation with Dynamical Systems, Physica D 120:120--214, 1997.

\bibitem{Sloman}
Sloman, A., Afterthoughts on analogical representations, in: R. Brachman, H. Levesque (Eds.), Readings in Knowledge Representation, Morgan and Kaufmann, Los Altos, CA, pp. 431--440, 1985.

\bibitem{Sun}
Sun, G. Z., Chen, H. H.,  and Lee, Y. C., Turing equivalence of neural networks with second order connection weights, IJCNN-91-Seattle International Joint Conference on Neural Networks, Seattle, WA, pp. 357-362 vol.2., 1991. doi: 10.1109/IJCNN.1991.155360

\bibitem{Brian-Smith}
Smith, B. C., Prologue to ``Reflection and Semantics in a Procedural Language'', in: R. Brachman, H. Levesque (Eds.), Readings in Knowledge Representation, Morgan and Kaufmann, Los Altos, CA, 1985, pp. 31-40.

%\bibitem{Stapleton}
%Stapleton, G. A Survey of Reasoning Systems Based on Euler Diagrams, Electronic Notes in Theoretical Computer Science, 134 (2005), pp. 127-151.

%\bibitem{Stiny}
%Stiny, G., Shape: Talking about Seeing and Doing. MIT Press, Cambridge, MA, 2006.

\bibitem{TuringMachine}
Turing, A.M. On Computable Numbers, with an Application to the Entscheidungs problem, Proceedings of the London Mathematical Society, 2 (published 1937), 42:230--265, 1936.

\bibitem{Turing}
Turing, A., Computing machinery and Intelligence, Mind 59:433--460, 1950.

\bibitem{Tye}
Tye, M., The Imagery Debate, A Bradford Book, The MIT Press, Cambridge, Mass., 1991.

%\bibitem{Wittgenstein}
%Wittgenstein, L., Philosophical Investigations, Basil Blackwell, Oxford, 1953.

\end{thebibliography}

\end{document}